\pdfoutput=1

\documentclass[11pt]{article}

\usepackage[]{acl}

\usepackage{times}
\usepackage{latexsym}

\usepackage{placeins}
\usepackage{multirow}
\usepackage{graphicx}
\usepackage{pgf}
\usepackage{layouts}
\usepackage{hyperref}
\usepackage{mdframed}
\usepackage{lipsum}
\usepackage{paralist}
\usepackage{soul}

\usepackage[T1]{fontenc}

\usepackage[utf8]{inputenc}

\usepackage{microtype}

\usepackage{booktabs}
\usepackage{arydshln}
\usepackage{titlesec}
\titlespacing{\paragraph}{%
  0pt}{
  0.2\baselineskip}{
  1em}%

\title{Pre-trained Language Models for Keyphrase Generation:\\ A Thorough Empirical Study}

\author{
Di Wu, Wasi Uddin Ahmad, Kai-Wei Chang \\
University of California, Los Angeles \\ 
\texttt{\{diwu,kwchang\}@cs.ucla.edu, wasiahmad@ucla.edu}
}

\begin{document}
\maketitle

\begin{abstract}
    Neural models that do not rely on pre-training have excelled in the keyphrase generation task with large annotated datasets. Meanwhile, new approaches have incorporated pre-trained language models (PLMs) for their data efficiency. However, there lacks a systematic study of how the two types of approaches compare and how different design choices can affect the performance of PLM-based models. To fill in this knowledge gap and facilitate a more informed use of PLMs for keyphrase extraction and keyphrase generation, we present an in-depth empirical study. Formulating keyphrase extraction as sequence labeling and keyphrase generation as sequence-to-sequence generation, we perform extensive experiments in three domains. After showing that PLMs have competitive high-resource performance and state-of-the-art low-resource performance, we investigate important design choices including in-domain PLMs, PLMs with different pre-training objectives, using PLMs with a parameter budget, and different formulations for present keyphrases. Further results show that (1) in-domain BERT-like PLMs can be used to build strong and data-efficient keyphrase generation models; (2) with a fixed parameter budget, prioritizing model depth over width and allocating more layers in the encoder leads to better encoder-decoder models; and (3) introducing four in-domain PLMs, we achieve a competitive performance in the news domain and the state-of-the-art performance in the scientific domain.
\end{abstract}
\section{Introduction}
Keyphrases are phrases that condense salient information of a document. 
Because of their high information density, keyphrases have been widely used for indexing documents, linking to relevant information, or recommending products \citep{10.1145/1367497.1367723, 10.1145/1871437.1871754}. Keyphrases have also functioned as important features for text summarization \citep{10.5555/1039791.1039794}, information retrieval \citep{10.1145/312624.312671, 10.1145/1141753.1141800, kim-etal-2013-applying, Tang2017QALinkET, boudin-etal-2020-keyphrase}, document clustering \citep{Hammouda2005}, and text classification \citep{10.3115/1220175.1220243, wilson-etal-2005-recognizing, berend-2011-opinion}. Given a document, a keyphrase is a \textit{present keyphrase} if it appears as a span in the document, or an \textit{absent keyphrase} otherwise. The task \textit{keyphrase extraction} requires a model to extract present keyphrases. \citet{meng-etal-2017-deep} introduce \textit{keyphrase generation}, where the model is required to predict all the present and absent keyphrases.

\begin{figure}[!t]
    \centering
    \small
    \resizebox{\linewidth}{!}{
    \begin{tabular}{ p{0.98\linewidth} }
    \toprule \\
    [-2ex]
    \textbf{Document title} \\
    \hdashline \\
    [-2ex]
    J.F.K. Workers Moved Drugs, Authorities Say \\
    \midrule
    \textbf{Document body} \\
    \hdashline \\
    [-2ex]
    Airline employees exploited weaknesses in security procedures to help a New York drug ring \textcolor{blue}{smuggle} \textcolor{blue}{heroin} and cocaine through \textcolor{blue}{Kennedy International Airport}, federal authorities charged yesterday. ... \\
    \midrule \\ [-2ex]
    \textbf{Present and Absent Keyphrases} \\
    \hdashline \\
    [-2ex]
    \textcolor{blue}{smuggling}, \textcolor{blue}{heroin}, \textcolor{blue}{kennedy international airport}, \textcolor{red}{drug abuse and traffic}, \textcolor{red}{crime and criminals}, \textcolor{red}{cocaine and crack cocaine} \\
    \bottomrule
    \end{tabular}}
    \caption{
    An example news article with present and absent keyphrases highlighted in blue and red respectively. For better readability, the document is not tokenized.
    }
    \label{example-case}
    \vspace{-2mm}
\end{figure}

With the successful application of pre-trained language models (PLMs) on various NLP tasks \citep{devlin-etal-2019-bert,brown2020language,lewis-etal-2020-bart,JMLR:v21:20-074,conneau-etal-2020-unsupervised}, PLMs have started to be incorporated in keyphrase extraction and generation methods. For instance, PLMs have been used for unsupervised keyphrase extraction \citep{8954611,liang-etal-2021-unsupervised}, keyphrase extraction via sequence labeling \citep{arxiv.1910.08840,9481005}, and keyphrase generation via sequence-to-sequence (seq2seq) generation  \citep{9443960,arxiv.2004.10462,2201.05302,kulkarni-etal-2022-learning,gao-etal-2022-retrieval,arxiv.2203.08118}. Nevertheless, these studies remain at an application level, without carefully evaluating the advantage or limitations of PLMs, or the effect of various decision choices when using PLMs. The shortage of established results leads to a range of high-level questions in the keyphrase generation research community: \textit{When should we use PLMs? Are we using the right PLMs? What PLMs should we use when there are constraints on annotated data or computational resources?} As prior works have already shown the advantage of PLMs in zero-shot \citep{kulkarni-etal-2022-learning}, multilingual \citep{gao-etal-2022-retrieval}, and low-resource \citep{arxiv.2203.08118} keyphrase generation, it is imperative to fill in this knowledge gap to provide future research on PLM-based keyphrase generation with a holistic view. 

In this paper, we carefully investigate the strategy for using PLMs for keyphrase extraction and generation, with emphases on four important dimensions. (1) \textbf{Encoder-only vs. encoder-decoder PLMs}. Following the seq2seq formulation in \citet{yuan-etal-2020-one}, previous studies often use encoder-decoder PLMs \footnote{We use seq2seq PLMs and encoder-decoder PLMs interchangeably in this paper.} such as BART \citep{lewis-etal-2020-bart} or T5 \citep{JMLR:v21:20-074} for keyphrase generation. However, the masked language modeling pre-training used by BERT-like encoder-only PLMs \citep{devlin-etal-2019-bert} may be inherently close to keyphrase generation that requires identifying short phrases that may be absent from the input. (2) \textbf{The pre-training domain}. Motivated by the availability of a large amount of in-domain BERT-like PLMs (Figure \ref{various-bert-models}), we investigate the performance gap between in-domain PLMs, out-of-domain PLMs, and general-domain PLMs. We further introduce four PLMs pre-trained using S2ORC \citep{lo-etal-2020-s2orc} or RealNews \citep{zellers2019neuralfakenews} and compare between BERT-like and BART-like in-domain PLMs. (3) \textbf{Parameter allocation strategies} with a constrained budget. To build a seq2seq keyphrase generation model with a limited parameter budget, is it better to make the model deeper, or wider? Moreover, should we allocate more parameters to the encoder, or the decoder? We investigate this problem using BERT with different sizes from \citet{arxiv.1908.08962}. (4) \textbf{Extraction vs. generation formulation}. For a given PLM, we compare three objectives for present keyphrases: generation, sequence labeling, and sequence labeling with Conditional Random Field \citep{10.5555/645530.655813}.

We conduct extensive experiments on seven datasets across three domains. The main findings are summarized as follows.

\begin{compactenum}
    \item  We show that large or in-domain seq2seq PLMs approach the performance of prior state-of-the-art keyphrase generation methods while being much more \textbf{sample efficient}.
    \item For the first time, we show that directly fine-tuning in-domain \textbf{encoder-only PLMs} gives strong keyphrase generation performance.
    \item With a limited parameter budget, we find that the \textbf{model depth} should be prioritized. Moreover, using a \textbf{deep encoder} with a \textbf{shallow decoder} greatly outperforms the reverse for keyphrase quality and inference latency.
    \item We systematically compare three different objectives for identifying present keyphrases and make concrete recommendations.
    \item We pre-train in-domain seq2seq PLMs \textbf{SciBART} and \textbf{NewsBART}. SciBART-large adapted to OAGKX \citep{cano-bojar-2020-two} outperforms KeyBART \citep{kulkarni-etal-2022-learning} and establishes the state-of-the-art scientific keyphrase generation performance.
\end{compactenum}

We hope this empirical study can shed light on the challenges and opportunities of using PLMs for keyphrase generation and motivate the development of more effective and efficient approaches. The code and pre-trained models will be publicly available at \url{https://github.com/uclanlp/DeepKPG}.
\section{Methods}
This section formulates keyphrase extraction and keyphrase generation in 2.1 and introduces the considered PLM-based approaches in 2.2 and 2.3.

\subsection{Problem Definition}
We view a keyphrase example as a triple \textbf{$(\mathbf{x},\mathbf{y_p},\mathbf{y_a})$}, corresponding to the input document $\mathbf{x}=(x_1,x_2,...,x_d)$, the set of present keyphrases $\mathbf{y_p}=\{y_{p_1}, y_{p_2}, ..., y_{p_m}\}$, and the set of absent keyphrases $\mathbf{y_a}=\{y_{a_1}, y_{a_2}, ..., y_{a_n}\}$. For both keyphrase extraction and generation, $\mathbf{x}$ consists of the title and the document body, concatenated with a special $[sep]$ token. Following \citet{meng-etal-2017-deep}, each $y_{p_i}\in\mathbf{y_p}$ is a substring of $\mathbf{x}$, and each $y_{a_i}\in\mathbf{y_a}$ does not appear in $\mathbf{x}$.

Using this formulation, the \textbf{keyphrase extraction} task requires the model to predict $\mathbf{y_p}$. In this paper, we use a sequence labeling formulation. Each $x_i\in \mathbf{x}$ is assigned a label $c_i\in\{B_{kp}, I_{kp}, O\}$ depending on $x$ being the beginning token of a present keyphrase, the subsequent token of a present keyphrase, or otherwise. The model is required to predict the label for each token. On the other hand, the \textbf{keyphrase generation} task requires the prediction of $\mathbf{y_p} \cup \mathbf{y_a}$. Following \citet{yuan-etal-2020-one}, we use a special separator token \texttt{;} to join all the keyphrases as the target sequence $\mathbf{y}=(y_{p_1}$ \texttt{;} $...$  \texttt{;} $y_{p_m}$ \texttt{;} $y_{a_1}$ \texttt{;} $...$  \texttt{;} $y_{a_m})$.

\subsection{Keyphrase Extraction}
For keyphrase extraction, we fine-tune four encoder-only PLMs: \textbf{BERT} \citep{devlin-etal-2019-bert}, \textbf{RoBERTa} \citep{arxiv.1907.11692}, \textbf{SciBERT} \citep{beltagy-etal-2019-scibert}, and \textbf{NewsBERT} (section \ref{section_scibart}) \footnote{In this study, we use the base variants of all the encoder-only models unless otherwise specified.}. We add a fully connected layer for each model to project the hidden representation of every token into three logits representing $B_{kp}$, $I_{kp}$, and $O$. The model is trained on the cross-entropy loss. We also experiment with using Conditional Random Field \citep{10.5555/645530.655813} to model the sequence-level transitions better. We use \textbf{+CRF} to refer to the models with this change.

\subsection{Keyphrase Generation}
\subsubsection{Encoder-Decoder PLMs}
\label{subsubsec:seq2seq_plm}
Using the sequence generation formulation above, we fine-tune \textbf{BART} \citep{lewis-etal-2020-bart}, \textbf{T5} \citep{JMLR:v21:20-074}, \textbf{SciBART} (section \ref{section_scibart}), and \textbf{NewsBART} (section \ref{section_scibart}). The models are trained with cross-entropy loss for generating the target sequence of concatenated keyphrases. 

\subsubsection{Encoder-only PLMs}
\label{subsubsec:enc_only_plm}
\paragraph{BERT2BERT}
We construct seq2seq keyphrase generation models by separately initializing the encoder and the decoder with encoder-only PLMs. Following \citet{rothe-etal-2020-leveraging}, we add cross-attention layers to the decoder. The model is fine-tuned as in \ref{subsubsec:seq2seq_plm}. We use five pre-trained BERT checkpoints from \citet{arxiv.1908.08962} with hidden size 768, 12 attention heads per layer, and 2, 4, 6, 8, and 10 layers. \textbf{B2B-$e$+$d$} denotes a BERT2BERT model with an $e$-layer pre-trained BERT as the encoder and a $d$-layer pre-trained BERT as the decoder. We use BERT2RND (\textbf{B2R}) to denote randomly initializing the decoder and RND2BERT (\textbf{R2B}) to denote randomly initializing the encoder. 

\paragraph{Mask Manipulation}
\citet{arxiv.1905.03197} propose jointly pre-training for unidirectional, bidirectional, and seq2seq language modeling by controlling mask patterns. In the seq2seq setup, the input is $\mathbf{x}\ [eos]\ \mathbf{y}$. The attention mask is designed such that tokens in $\mathbf{x}$ are only allowed to attend to other tokens within $\mathbf{x}$, and that tokens in $\mathbf{y}$ are only allowed to attend to tokens on their left. Using this formulation, we fine-tune encoder-only PLMs for seq2seq keyphrase generation. Following \citet{arxiv.1905.03197}, we mask and randomly replace tokens from $\mathbf{y}$ and train the model on the cross-entropy loss between its reconstruction and the original sequence. We call our models \textbf{BERT-G}, \textbf{RoBERTa-G}, \textbf{SciBERT-G}, and \textbf{NewsBERT-G}. For a more comprehensive comparison, we also experiment on the UniLMv2 model without relative position bias \citep{arxiv.2002.12804}, denoted as \textbf{UniLM}.

\subsubsection{Domain-specific PLMs}
\label{section_scibart}
Previous works have established the advantage of domain-specific PLMs in a wide range of tasks  \citep{beltagy-etal-2019-scibert, gururangan-etal-2020-dont}. To fill the vacancy of encoder-decoder PLMs in the scientific and news domain, we pre-train \textbf{SciBART} and \textbf{NewsBART}. The pre-training details are presented in the appendix section \ref{appendix_plm_pretraining}.

\paragraph{Scientific PLMs} We pre-train \textbf{SciBART-base} and \textbf{SciBART-large} using all the paper titles and abstracts from the S2ORC dataset \citep{lo-etal-2020-s2orc}. The dataset contains 171.7M documents or 15.4B tokens in total. Following \citet{beltagy-etal-2019-scibert}, we construct a scientific BPE vocabulary based on the pre-training data. The model is pre-trained using text infilling for 250k steps with a batch size of 2048 and other BART pertaining hyperparameters. 

\paragraph{News PLMs} We pre-train \textbf{NewsBART-base} using the RealNews dataset \citep{zellers2019neuralfakenews}, which contains around 130GB of news text from 2016 to 2019. We start from the BART-base checkpoint and pre-train on text infilling for 250k steps with batch size 2048. To enable comparisons with SciBERT, we also pre-train a \textbf{NewsBERT} model using the same data on masked language modeling starting from BERT-base \citep{devlin-etal-2019-bert} for 250k steps with a batch size 512. 

\section{Experimental Setup}

\subsection{Benchmarks}
We test the methods in three domains: science, news, and online forum. The statistics of the testing datasets are provided in the appendix section \ref{appendix-data-stats}.

\paragraph{\textbf{SciKP}} \citet{meng-etal-2017-deep} introduce KP20k, a large keyphrase generation dataset containing 500k Computer Science papers. Following their work, we train on KP20k and evaluate on the KP20k test set and four out-of-distribution testing datasets: Inspec \citep{10.3115/1119355.1119383}, Krapivin \citep{Krapivin2009LargeDF}, NUS \citep{10.1007/978-3-540-77094-7_41}, and SemEval \citep{kim-etal-2010-semeval}.

\paragraph{\textbf{KPTimes}} Introduced by \citet{gallina-etal-2019-kptimes}, KPTimes is a keyphrase generation dataset in the news domain containing over 250k examples. We train on the KPTimes train set and report the performance on the union of the KPTimes test set and the out-of-distribution test set JPTimes.

\paragraph{\textbf{StackEx}} We also establish the performance of various PLMs on the StackEx dataset \citep{yuan-etal-2020-one}. The results are presented in the appendix.

\subsection{Baselines}
We compare PLMs with four supervised keyphrase generation baselines.
\textbf{CatSeq} \citep{yuan-etal-2020-one} is a CopyRNN \citep{meng-etal-2017-deep} trained on generating keyphrases as a sequence, separated by the separator token.
\textbf{ExHiRD-h} \citep{chen-etal-2021-training} is an improved version of CatSeq, where a hierarchical decoding framework and a hard exclusion mechanism are used to reduce duplicates.
\textbf{Transformer} is the self-attention based seq2seq model \citep{vaswani2017attention} with copy mechanism, while \textbf{SetTrans} \citet{ye-etal-2021-one2set} performs order-agnostic keyphrase generation. The model is trained via a k-step target assignment algorithm.

For keyphrase extraction, we include a range of traditional keyphrase extraction methods and a randomly initialized \textbf{Transformer} as the baselines. Appendix section \ref{appendix_baselines} introduces all baselines and their implementation details. 

\subsection{Evaluation}
Each method's predictions are normalized into a sequence of present and absent keyphrases. The phrases are ordered by the position in the source document for the sequence labeling approaches to obtain the keyphrase predictions. Then, we apply the Porter Stemmer \citep{Porter1980AnAF} to the output and target phrases and remove the duplicated phrases from the output. Following \citet{chan-etal-2019-neural} and \citet{chen-etal-2020-exclusive}, we report the macro-averaged F1@5 and F1@M scores. For all the results except the ablation studies, we train with three different random seeds and report the averaged scores. 

\subsection{Implementation Details}
We provide the full implementation details in appendix section \ref{appendix_plm_implementations} and hyperparameter details in appendix section \ref{appendix-hyperparams}. For evaluation, we follow \citet{chan-etal-2019-neural}'s implementation.
\section{Results}
We aim to address the following questions.
\begin{compactenum}
    \item How do PLMs perform on keyphrase generation compared to prior state-of-the-art non-PLM methods? Do PLMs help in the low-resource setting?
    \item How does the pre-training domain affect keyphrase generation performance?
    \item Can encoder-only PLMs generate better keyphrases than encoder-decoder PLMs?
    \item What is the best parameter allocation strategy for using encoder-decoder PLMs to balance the performance and computational cost?
\end{compactenum}

\setlength{\tabcolsep}{3.5pt}
\begin{table}[]
    \small
    \centering
    \resizebox{\linewidth}{!} {%
    \begin{tabular}{l | c | c c | c c  }
    \hline
    \multirow{2}{*}{Method} &
    \multirow{2}{*}{|M|} & \multicolumn{2}{c|}{\textbf{KP20k}} & \multicolumn{2}{c}{\textbf{KPTimes}} \\
    & & F1@5 & F1@M & F1@5 & F1@M \\
    \hline
    \multicolumn{4}{l}{\textbf{Present keyphrase generation}} \\ 
    \hline
    CatSeq & 21M & 29.1 & 36.7 & 29.5 & 45.3 \\
    ExHiRD-h & 22M & 31.1 & 37.4 & 32.1 & 45.2 \\
    Transformer & 98M & 33.3 & 37.6 & 30.2 & 45.3 \\
    SetTrans & 98M & \textbf{35.6} & 39.1 & 35.6 & 46.3 \\
    \hdashline
    \multirow{2}{*}{BART} & 140M & 32.2 & 38.8 & 35.9 & 49.9 \\
     & 406M & 33.2 & 39.2 & \textbf{37.3} & \textbf{51.0} \\
    \hdashline
    \multirow{2}{*}{T5} & 223M & 33.6 & 38.8 & 34.6 & 49.2 \\
     & 770M & 34.3 & \textbf{39.3} & 36.6 & 50.8 \\
    \hline
    \multicolumn{4}{l}{\textbf{Absent keyphrase generation}} \\ 
    \hline
    CatSeq & 21M & 1.5 & 3.2 & 15.7 & 22.7 \\
    ExHiRD-h & 22M & 1.6 & 2.5 & 13.4 & 16.5 \\
    Transformer & 98M & 2.2 & 4.6 & 17.1 & 23.1 \\
    SetTrans & 98M & \textbf{3.5} & \textbf{5.8} & \textbf{19.8} & 21.9 \\
    \hdashline
    \multirow{2}{*}{BART} & 140M & 2.2 & 4.2 & 17.1 & \textbf{24.9} \\
     & 406M & 2.7 & 4.7 & 17.6 & 24.4 \\
    \hdashline
    \multirow{2}{*}{T5} & 223M & 1.7 & 3.4 & 15.3 & 24.2 \\
     & 770M & 1.7 & 3.5 & 15.7 & 24.1 \\ 
    \hline
    \end{tabular}
    }
    \caption{
    Keyphrase generation performance of state-of-the-art non-PLM methods and seq2seq PLMs. The best results are boldfaced. Large-sized PLMs approach the performance of SetTrans on KP20k and obtain the state-of-the-art F1@M on KPTimes.
    }
    \label{tab:main-kpg-results}
\end{table}

\begin{figure}[]
\centering
\vspace{-8mm}
\includegraphics[width=0.48\textwidth]{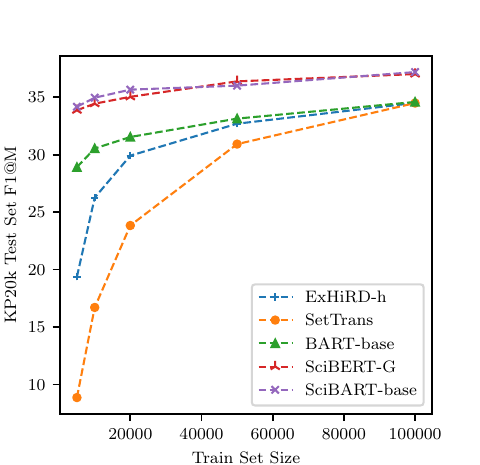}
\vspace{-6mm}
\caption{Present keyphrase generation performance of different methods as a function of training set size.}
\vspace{-2mm}
\label{perf-vs-resource-kp20k}
\end{figure}

\subsection{A gentle start: seq2seq PLMs vs. SOTA non-PLM keyphrase generation methods}
We compare the present and absent keyphrase generation performance of four popular seq2seq PLMs with the performance of state-of-the-art non-PLM keyphrase generation approaches. The results are presented in Table \ref{tab:main-kpg-results} and Figure \ref{perf-vs-resource-kp20k}. 

\paragraph{Large PLMs have competitive performance in keyphrase generation benchmarks.} 
According to Table \ref{tab:main-kpg-results}, BART and T5 approach the performance of SetTrans on SciKP and outperform the state-of-the-art on KPTimes. Out-of-distribution SciKP test sets also exhibit the same pattern (Table \ref{tab:scikp-all-results-pkp} and \ref{tab:scikp-all-results-akp} in the appendix). Moreover, scaling up the size of PLMs from base to large greatly improves present keyphrase generation but has a relatively small effect on absent keyphrase generation. 

\paragraph{PLMs have an absolute advantage in low-resource keyphrase generation.}
Due to the knowledge accumulated during pre-training, PLMs are often observed as data-efficient learners on various downstream NLP tasks. We verify this pattern in keyphrase generation by training PLMs and non-PLM methods using a small training set consisting of 5k, 10k, 20k, 50k, and 100k examples. From Figure \ref{perf-vs-resource-kp20k}, we observe that PLMs require only 20k annotated KP20k training examples to perform, as good as non-PLM methods trained with 100k examples. Figure \ref{perf-vs-resource-kptimes} in the appendix confirms a similar trend in the news domain using KPTimes.

To summarize, this section establishes that (1) seq2seq PLMs can approach SOTA keyphrase generation performance and (2) in the low-resource regime, PLMs are very data-efficient and greatly outperform SOTA methods trained from scratch. We emphasize that this comparison aims at faithfully establishing the performance of fine-tuning commonly used seq2seq PLMs for keyphrase generation with the sequence generation formulation. To build more advanced models, future work may consider combining SetTrans with PLMs.

\subsection{Is keyphrase generation performance sensitive to the domain of PLMs?}
In this section, we establish that the pre-training domain greatly affects the keyphrase generation performance of seq2seq PLMs.

\paragraph{In-domain PLMs benefit rich and low resource keyphrase generation.}
Table \ref{tab:non_original_seq2seq} compares the performance of SciBART and NewsBART with BART. We find that SciBART greatly improves both present and absent keyphrase generation performance on KP20k, while NewsBART brings large gains of absent keyphrase performance on KPTimes. In low-resource regimes, the in-domain SciBART greatly outperforms the general-domain BART-base on KP20k (Figure \ref{perf-vs-resource-kp20k}). The strong performance in all resource settings indicates the importance of using domain-specific PLMs. 

\setlength{\tabcolsep}{3.5pt}
\begin{table}[!t]
    \small
    \centering
    \resizebox{\linewidth}{!} {%
    \begin{tabular}{l | c | c c | c c  }
    \hline
    \multirow{2}{*}{Method} &
    \multirow{2}{*}{|M|} & \multicolumn{2}{c|}{\textcolor{teal}{\textbf{KP20k}}} & \multicolumn{2}{c}{\textcolor{violet}{\textbf{KPTimes}}} \\
    & & F1@5 & F1@M & F1@5 & F1@M \\
    \hline
    \multicolumn{4}{l}{\textbf{Present keyphrase generation}} \\ 
    \hline
    \multicolumn{6}{l}{Encoder-only PLM} \\ 
    \hdashline
    BERT-G & 110M & 31.3 & 37.9 & 32.3 & 47.4 \\
    B2B-8+4 & 143M & 32.2 & 38.0 & \underline{33.8} & \underline{48.6} \\
    RoBERTa-G & 125M & 28.8 & 36.9 & 33.0 & 48.2 \\
    \textcolor{teal}{SciBERT-G} & \textcolor{teal}{110M} & \textcolor{teal}{\underline{32.8}} & \textcolor{teal}{\underline{\textbf{39.7}}} & \textcolor{teal}{33.0} & \textcolor{teal}{48.4} \\    
    \textcolor{violet}{NewsBERT-G} & \textcolor{violet}{110M} &\textcolor{violet}{29.9} & \textcolor{violet}{36.8} & \textcolor{violet}{33.0} & \textcolor{violet}{48.0} \\   
    \hdashline
    \multicolumn{6}{l}{Encoder-Decoder PLM} \\ 
    \hdashline
    BART & 140M & 32.2 & 38.8 & \textbf{35.9} & \textbf{49.9} \\
    \textcolor{teal}{SciBART} & \textcolor{teal}{124M} & \textcolor{teal}{\textbf{34.1}} & \textcolor{teal}{39.6} & \textcolor{teal}{34.8} & \textcolor{teal}{48.8} \\
    \textcolor{violet}{NewsBART} & \textcolor{violet}{140M} & \textcolor{violet}{32.4} & \textcolor{violet}{38.7} & \textcolor{violet}{35.4} & \textcolor{violet}{49.8} \\
    \hline
    \multicolumn{4}{l}{\textbf{Absent keyphrase generation}} \\ 
    \hline
    \multicolumn{6}{l}{Encoder-only PLM} \\ 
    \hdashline
    BERT-G & 110M & 1.9 & 3.7 & 16.5 & 24.6 \\
    B2B-8+4 & 143M & 2.2 & 4.2 & 16.8 & 24.5 \\
    RoBERTa-G & 125M & 2.0 & 3.9 & \underline{17.1} & 25.5 \\
    \textcolor{teal}{SciBERT-G} & \textcolor{teal}{110M} & \textcolor{teal}{\underline{2.4}} & \textcolor{teal}{\underline{4.6}} & \textcolor{teal}{15.7} & \textcolor{teal}{24.7} \\
    \textcolor{violet}{NewsBERT-G} & \textcolor{violet}{110M} & \textcolor{violet}{1.3} & \textcolor{violet}{2.6} & \textcolor{violet}{17.0} & \textcolor{violet}{\underline{25.6}} \\ 
    \hdashline
    \multicolumn{6}{l}{Encoder-Decoder PLM} \\ \hdashline
    BART & 140M & 2.2 & 4.2 & 17.1 & 24.9 \\
    \textcolor{teal}{SciBART} & \textcolor{teal}{124M} & \textcolor{teal}{\textbf{2.9}} & \textcolor{teal}{\textbf{5.2}} & \textcolor{teal}{17.2} & \textcolor{teal}{24.6}  \\
    \textcolor{violet}{NewsBART} & \textcolor{violet}{140M} & \textcolor{violet}{2.2} & \textcolor{violet}{4.4} & \textcolor{violet}{\textbf{17.6}} & \textcolor{violet}{\textbf{26.1}} \\
    \hline
    \end{tabular}
    }
    \vspace{-2mm}
    \caption{
    A comparison across encoder-only and encoder-decoder PLMs from different domains for keyphrase generation. The best results are boldfaced, and the best encoder-only PLM results are underlined. We use different colors for PLMs in the \textcolor{teal}{science} and the \textcolor{violet}{news} domain.
    }
    \label{tab:non_original_seq2seq}
    \vspace{-2mm}
\end{table}

\begin{figure}[!t]
\centering
\includegraphics[width=0.48\textwidth]{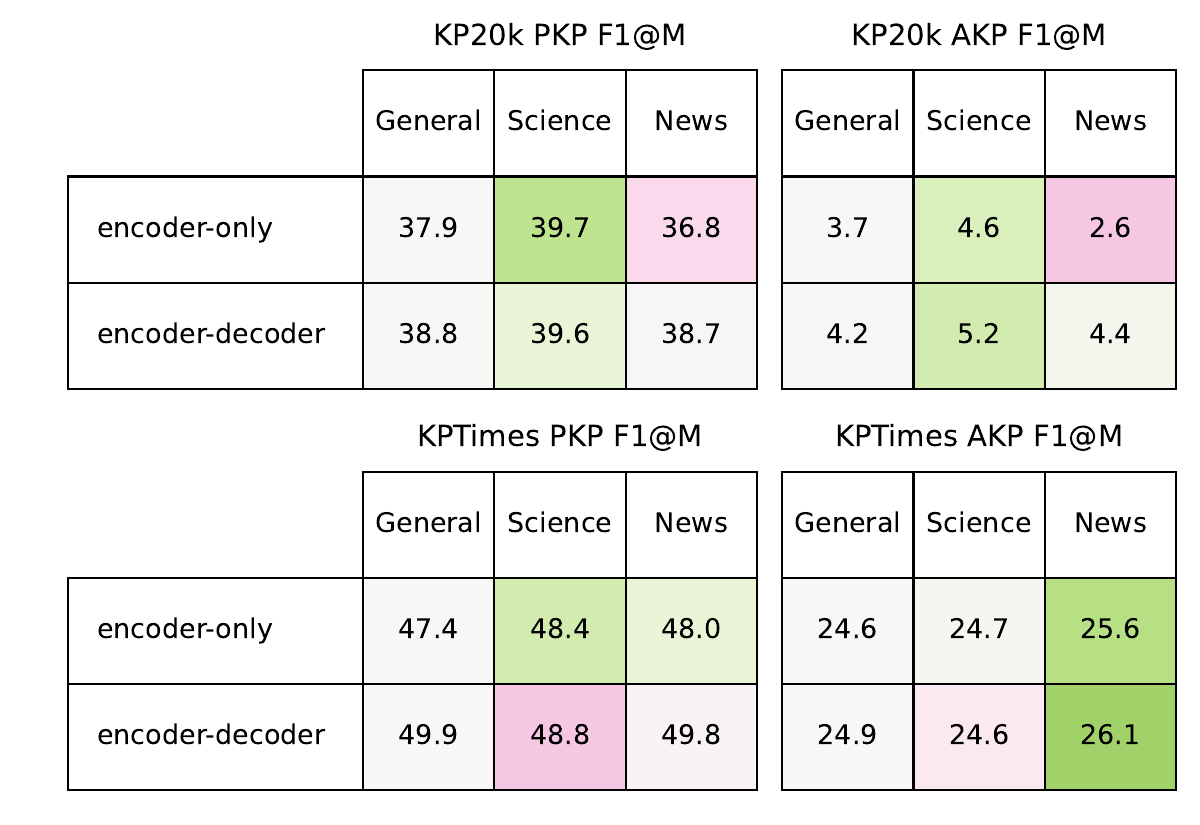}
\vspace{-6mm}
\caption{Keyphrase generation performance of PLMs pre-trained in different domains. Each table shows the performance of BERT-G, SciBERT-G, and NewsBERT-G in the first row and the performance of BART, SciBART, and NewsBART in the second row.}
\vspace{-2mm}
\label{perf-vs-plm-domain}
\end{figure}

\paragraph{Domain sensitivity map} In Figure \ref{perf-vs-plm-domain}, we investigate the sensitivity of the performance to the domain of PLMs. We compare BERT-G, SciBERT-G, NewsBERT-G, BART-base, SciBART-base, and NewsBART-base, representing encoder-only and encoder-decoder models from three distinct domains. In 7 out of 8 comparisons, in-domain PLMs outperform domain-general PLMs; in 5 out of 8, out-of-domain PLMs perform worse. Furthermore, \textbf{absent keyphrases} are more sensitive to the pre-training domain, and that science PLMs have good transferability to the news domain, while the reverse is not true.

\subsection{Can encoder-only PLMs generate better keyphrases than encoder-decoder PLMs?}

In this section, we shift the focus from seq2seq PLMs to encoder-only PLMs and compare (1) training for sequence generation by manipulating masks (BERT-G, RoBERTa-G, SciBERT-G, and NewsBERT-G), (2) the best performance of BERT2BERT models with a 12-layer budget (full results are discussed in the next section), and (3) BART-based seq2seq PLMs. Table \ref{tab:non_original_seq2seq} presents the results on KP20k and KPTimes.

We start with the surprising result that strong keyphrase generation models can be obtained using seq2seq attention masks to fine-tune encoder-only PLMs. On KP20k, SciBERT-G outperforms BART on all the metrics. On KPTimes, NewsBERT-G has comparable F1@5 and better F1@M for absent keyphrase generation compared to BART. Overall, our results have the important implication that \textbf{in the absence of in-domain seq2seq PLMs, an in-domain encoder-only PLM may be preferred over the domain-general BART for keyphrase generation}. This preference is more evident in low-resource scenarios (Figure \ref{perf-vs-resource-kp20k}). SciBERT only requires 5k data to achieve the same F1@M of BART fine-tuned with 100k data. 

Next, we observe that combining two smaller-sized BERT models and training on keyphrase generation produce better results than BERT-G despite having a similar model size. On KPTimes, the B2B model with an 8-layer encoder and a 4-layer decoder achieves the best present keyphrase performance among all encoder-only PLMs. The model also has \textbf{a lower inference latency} due to its shallow decoder structure (appendix section \ref{appendix-inference-speed}).

To summarize, this section shows that in-domain encoder-only PLMs may achieve better keyphrase generation performance than domain-general encoder-decoder PLMs. They also enjoy better resource efficiency.

\subsection{What is the best parameter allocation strategy for encoder-decoder models?}

\setlength{\tabcolsep}{3.5pt}
\begin{table}[!t]
    \small
    \centering
    \begin{tabular}{l | c | c c | c c }
    \hline
    \multirow{2}{*}{Model setup} & \multirow{2}{*}{|M|} & \multicolumn{2}{c|}{\textbf{Present KPs}} & \multicolumn{2}{c}{\textbf{Absent KPs}} \\
    & & F1@5 & F1@M & F1@5 & F1@M \\
    \hline
    B2B-12+12-128 & 13M & \textbf{26.4} & \textbf{33.8} & \textbf{1.0} & \textbf{2.1} \\
    B2B-2+2-256 & 20M & 22.1 & 31.9 & \textbf{1.0} & \textbf{2.1} \\
    \hline
    B2B-12+12-256 & 38M & \textbf{30.8 }& \textbf{36.7} & \textbf{1.6} & \textbf{3.3} \\
    B2B-2+2-512 & 47M & 27.5 & 34.7 & 1.4 & 3.0 \\
    \hline
    B2B-10+4-512 & 81M & \textbf{31.4} & \textbf{37.4} &\textbf{ 1.8 }& \textbf{3.7} \\
    B2B-2+2-768 & 82M & 29.4 & 35.5 &\textbf{ 1.8} & 3.5 \\
    \hline
    B2B-12+6-512 & 95M & \textbf{31.9} & \textbf{38.2} &\textbf{ 2.1} & \textbf{4.0} \\
    B2B-4+2-768 & 96M & 30.8 & 37.1 & 2.0 & \textbf{4.0} \\
    \hline
    \end{tabular}
    \caption{
    Comparison between different parameter allocation strategies. The best performance of each group is boldfaced. B2B-$e$+$d$-$h$ denotes a B2B model with $e$ encoder layers, $d$ encoder layers, and hidden size $h$.
    }
    \label{tab:depth-vs-width}
    \vspace{-2mm}
\end{table}

Observing that the BERT2BERT setup can produce strong keyphrase generation models, we further investigate the optimal parameter allocation. Given a parameter budget, should depth (i.e., more layers) or width (fewer layers, more parameters per layer) be prioritized? Moreover, should the encoder or the decoder be allocated more parameters? Computational resources prevent us from pre-training different models, so we use the BERT2BERT setting with different sized BERT models instead. 

\paragraph{Depth should be prioritized over width.}
We design four pairs of B2B models with different total parameter budgets: 20M, 50M, 85M, and 100M. Each pair contains (a) a model that prioritizes depth and (b) a model that prioritizes width. We make sure that (a) and (b) have similar encoder depth to decoder depth ratios (except group 3). The results on KP20k are presented in Table \ref{tab:depth-vs-width}. It is clear that model (a) performs significantly better for all the groups despite having slightly fewer parameters. 

\setlength{\tabcolsep}{3.5pt}
\begin{table}[!t]
    \small
    \centering
    \begin{tabular}{l | c | c | c c | c c  }
    \hline
    \multirow{2}{*}{$e$-$d$} &
    \multirow{2}{*}{|M|} & \multirow{2}{*}{Arch.} &   \multicolumn{2}{c|}{\textbf{KP20k}} & \multicolumn{2}{c}{\textbf{KPTimes}} \\
    & & & F1@5 & F1@M & F1@5 & F1@M \\
    \hline
    \multicolumn{6}{l}{\textbf{Present keyphrase generation}} \\ \hline
    2-10 & 158M & B2B & 30.4 & 36.4 & 31.6 & 46.5 \\ \hdashline
    \multirow{3}{*}{4-8} & \multirow{3}{*}{153M} & B2B & 31.7 & 37.7 & 32.9 & 47.6 \\
     &  & {R2B} & 26.3 & 35.2 & 28.2 & 43.3 \\
     &  & {B2R} & 31.7 & 37.9 & 32.6 & 47.5 \\ \hdashline
    \multirow{3}{*}{6-6} & \multirow{3}{*}{148M} & B2B & 32.1 & 37.7 & \textbf{33.8} & 48.4 \\
     &  & {R2B} & 26.4 & 35.3 & 27.8 & 42.9 \\
     &  & {B2R} & 32.0 & 38.4 & 33.3 & 48.2 \\ \hdashline
     \multirow{3}{*}{8-4} & \multirow{3}{*}{143M} & B2B & \textbf{32.2} & \textbf{38.0} & \textbf{33.8} & \textbf{48.6} \\
    &  & {R2B} & 27.3 & 35.4 & 27.8 & 42.8 \\
     &  & {B2R} & 31.2 & 37.9 & 33.2 & 48.0 \\ \hdashline
    10-2 & 139M & B2B & 31.7 & 38.0 & 33.5 & 48.4 \\
    \hline
    \multicolumn{6}{l}{\textbf{Absent keyphrase generation}} \\ \hline
    2-10 & 158M & B2B & 2.1 & 3.9 & 16.2 & 23.2 \\ \hdashline
    \multirow{3}{*}{4-8} & \multirow{3}{*}{153M} & B2B & \textbf{2.2} & 4.1 & 15.9 & 23.6 \\
    &  & {R2B} & 2.5 & 4.2 & 14.7 & 24.3 \\
    &  & {B2R} & 2.2 & 4.2 & 16.5 & 24.1 \\ \hdashline
    \multirow{3}{*}{6-6} & \multirow{3}{*}{148M} & B2B & \textbf{2.2} & 4.1 & 16.4 & 24.1 \\
    &  & {R2B} & 2.6 & 4.3 & 14.5 & 20.8 \\
    &  & {B2R} & 2.3 & 4.4 & 16.2 & 23.9 \\ \hdashline
    \multirow{3}{*}{8-4 } & \multirow{3}{*}{143M} & B2B & \textbf{2.2} & \textbf{4.2} & \textbf{16.8} & \textbf{24.5} \\
    &  & {R2B} & 2.4 & 4.1 & 14.9 & 21.0 \\
    &  & {B2R} & 2.1 & 4.1 & 16.8 & 24.7 \\ \hdashline
    10-2 & 139M & B2B & 2.1 & 4.1 & \textbf{16.8} & \textbf{24.5} \\
    \hline
    \end{tabular}
    \caption{
    A comparison between different BERT2BERT architectures. In $e$-$d$, $e$ and $d$ indicate the number of encoder and decoder layers, respectively. The best results among B2B models are boldfaced. 
    }
    \label{tab:b2b_ablations}
    \vspace{-2mm}
\end{table}

\paragraph{A deep encoder with a shallow decoder is preferred.} Next, we study the effect of layer allocation strategies. We fix a budget of 12 layers and experiment with five encoder-decoder combinations. Table \ref{tab:b2b_ablations} presents the results on KP20k and KPTimes. For both datasets, we find that the performance increases sharply and then plateaus as the depth of the encoder increases. With the same budget, \textbf{a deep encoder followed by a shallow decoder is strongly preferred over a shallow encoder followed by a deep decoder}. 
We hypothesize that comprehending the input article is important and challenging while generating a short string comprising several phrases based on the encoded article does not largely rely on the knowledge of PLMs. 

To verify, we conduct two further ablation studies by randomly initializing either the encoder ("R2B") or the decoder ("B2R"). The results are shown in Table \ref{tab:b2b_ablations}. For both datasets, we observe that randomly initializing the encoder greatly harms the performance, while randomly initializing the decoder does not significantly impact the performance (the absent keyphrase generation is even beneficial in some cases). 

In conclusion, with a limited parameter budget, we recommend using \textbf{more layers} and \textbf{a deep-encoder and shallow-decoder} architecture.
\section{Analysis}
In this section, we perform further analyses to investigate (1) different formulations for present keyphrase identification and (2) SciBART compared to KeyBART \citep{kulkarni-etal-2022-learning}. 

\subsection{Extraction vs. Generation: which is better for finding present keyphrases?}
Prior works have shown that PLMs with a generative formulation may improve the performance of information extraction tasks \citep{hsu-etal-2022-degree}. In Table \ref{tab:main-kpe-results}, we compare three formulations for identifying present keyphrases: (1) sequence labeling via token-wise classification, (2) sequence labeling with CRF, and (3) sequence generation\footnote{Results for all models are listed in the appendix.}. 

\setlength{\tabcolsep}{1pt}
\begin{table}[t]
    \small
    \centering
    \setlength{\tabcolsep}{2pt}
    \resizebox{\linewidth}{!} {%
    \begin{tabular}{l | c | c c | c c }
    \hline
    \multirow{2}{*}{Method} &
    \multirow{2}{*}{|M|} & \multicolumn{2}{c|}{\textbf{KP20k}} & \multicolumn{2}{c}{\textbf{KPTimes}} \\
    & & F1@5 & F1@M & F1@5 & F1@M \\
    \hline
    \multicolumn{6}{l}{\textbf{Present keyphrase extraction}} \\ \hline
    SciBERT & 110M & 28.6 & 40.5 & 31.8 & 47.7 \\
    NewsBERT & 110M & 25.8 & 37.5 & 34.5 & 50.4 \\
    \hdashline
    SciBERT+CRF & 110M & 28.4 & \textbf{42.1} & 31.8 & 48.1\\
    NewsBERT+CRF & 110M & 26.8 & 39.7 & \textbf{34.9} & \textbf{50.8} \\
    \hline
    \multicolumn{6}{l}{\textbf{Present keyphrase generation}} \\ \hline
    SciBERT-G & 110M & \textbf{32.8} & 39.7 & 33.0 & 48.4 \\ 
    NewsBERT-G & 110M & 29.9 & 36.8 & 33.0 & 48.0 \\   
    \hline
    \end{tabular}
    }
    \caption{
    Present keyphrase performance of PLM-based sequence labeling and sequence generation.
    }
    \label{tab:main-kpe-results}
    \vspace{-2mm}
\end{table}

For SciBERT and NewsBERT, we find that adding a CRF layer consistently improves the performance. Further comparing the results with (3), we find that the sequence labeling objective can guide the generation of more accurate (reflected by high F1@M) but fewer (reflected by low F1@5) keyphrases. Thus, for a given encoder-only PLM, the sequence labeling objective should be preferred if F1@M is important and generating absent keyphrases is not a concern. If generating absent keyphrases in a certain order is important, the sequence generation formulation should be preferred. However, if a strong in-domain seq2seq PLM is present, then the sequence generation should always be used (Table \ref{tab:scikp-all-results-pkp} and \ref{tab:other-all-results-pkp} in the appendix).

\subsection{Does task-specific pre-training waive the need for in-domain pre-training?}
KeyBART \citep{kulkarni-etal-2022-learning} is a recent approach of continued pre-training of BART using the OAGKX dataset \citep{cano-bojar-2020-two} on the keyphrase generation task with the keyphrases corrupted from the input text. On the other hand, SciBART only performs task-agnostic in-domain pre-training. To understand the effectiveness of these two training schemes, we fine-tune SciBART on keyphrase generation using OAGKX without corrupting the input text and evaluate the resulting model's zero-shot and transfer performance on KP20k. We use batch size 256, learning rate 3e-5, and 250k steps in total, which is approximately 2.8 epochs, comparable to \citet{kulkarni-etal-2022-learning} where the model is trained for two epochs.

\setlength{\tabcolsep}{3.5pt}
\begin{table}[!t]
    \small
    \centering
    \begin{tabular}{l | c | c c | c c  }
    \hline
    \multirow{2}{*}{Model} & \multirow{2}{*}{|M|} & \multicolumn{2}{c|}{\textbf{Present KPs}} & \multicolumn{2}{c}{\textbf{Absent KPs}} \\
    & & F1@5 & F1@M & F1@5 & F1@M \\
    \hline
    \multicolumn{6}{l}{\textbf{Zero-shot transfer}}\\
    \hline
    KeyBART & 406M & 20.4 & 22.8 & 1.7 & 0.9 \\
    \hdashline
    \multirow{2}{*}{SciBART$^\dagger$} & 124M & 26.6 & 31.2 & 1.5 & 2.6 \\
    & 386M & 23.5 & 28.5 & 1.3 & 2.3 \\
    \hline
    \multicolumn{6}{l}{\textbf{Fine-tuned on KP20k}}\\
    \hline
    KeyBART+ft & 406M & 32.5 & 39.8 & 2.6 & 4.7 \\ 
    \hdashline
    \multirow{2}{*}{SciBART+ft} & 124M & 34.1 & 39.6 & 2.9 & 5.2 \\
    & 386M & 34.7 & 41.5 & 3.1 & 5.7 \\
    \hdashline
    \multirow{2}{*}{SciBART$^\dagger$+ft} & 124M & 35.3 & 41.5 & 2.8 & 5.2 \\
    & 386M & \textbf{36.2} & \textbf{43.2} & \textbf{3.2} & \textbf{6.2} \\    
    \hline
    \end{tabular}
    \caption{
    Comparison between SciBART and KeyBART in zero-shot and fine-tuned settings. Both KeyBART and SciBART$^\dagger$ are first trained on OAGKX to learn to generate keyphrases. "+ft" means fine-tuned on KP20k. $^\dagger$ denotes the SciBART fine-tuned on OAGKX.
    }
    \label{tab:scibart_vs_keybart}
    \vspace{-2mm}
\end{table}

The results are presented in Table \ref{tab:scibart_vs_keybart}. Despite being 3x smaller, SciBART$^\dagger$-base outperforms KeyBART on the zero-shot transfer to KP20k. After fine-tuning on KP20k, SciBART$^\dagger$-base also has superior performance, while SciBART$^\dagger$-large establishes the state-of-the-art performance on KP20k. This suggests that \textbf{the gain from in-domain PLMs for keyphrase generation is fundamental even if a large-scale task-specific continued pre-training is performed}. 

\section{Related Work}

\paragraph{Keyphrase Extraction}
Early work on keyphrase extraction mainly followed a pipelined approach. First, a range of candidates (usually noun phrases) is selected by, e.g., regular expression matching. Then, various scoring methods are used to rank the candidates, and the ones with the highest scores are returned as keyphrase predictions \citep{hulth-2003-improved,mihalcea-tarau-2004-textrank,10.5555/1620163.1620205,bougouin-etal-2013-topicrank,8954611,boudin-2018-unsupervised, liang-etal-2021-unsupervised}. Later works adopt the sequence labeling formulation, which removes the need for selecting candidates \citep{zhang-etal-2016-keyphrase,luan-etal-2017-scientific,arxiv.1910.08840}. 

\paragraph{Keyphrase Generation}
\citet{meng-etal-2017-deep} propose the task of Deep Keyphrase Generation and a strong baseline model CopyRNN. The following works improve the architecture by adding correlation constraints \citep{chen-etal-2018-keyphrase} and linguistic constraints \citep{zhao-zhang-2019-incorporating}, exploiting learning signal from titles \citep{ye-wang-2018-semi, Chen2019TitleGuidedEF}, and hierarchical modeling the phrases and words \citep{chen-etal-2020-exclusive}. \citet{yuan-etal-2020-one} reformulate the problem as generating a sequence of keyphrases, while \citet{ye-etal-2021-one2set} further uses a set generation formulation to remove the influence of target phrase order. Other works include incorporating reinforcement learning \citep{chan-etal-2019-neural, luo-etal-2021-keyphrase-generation}, GANs \citep{swaminathan-etal-2020-preliminary}, and unifying keyphrase extraction with keyphrase generation \citep{chen-etal-2019-integrated,ahmad-etal-2021-select}. \citet{meng-etal-2021-empirical} conducts an empirical study on architecture, generalizability, phrase order, and decoding strategies.

More recently, \citet{arxiv.1910.08840}, \citet{9443960}, \citet{arxiv.2004.10462}, and \citet{9481005} have considered using pre-trained BERT \citep{devlin-etal-2019-bert} for keyphrase extraction and generation. In addition, \citet{2201.05302}, \citet{kulkarni-etal-2022-learning}, \citet{arxiv.2203.08118}, and \citet{gao-etal-2022-retrieval} use seq2seq PLMs such as BART or T5 in their approach. \citet{arxiv.2203.08118} show that domain-adaptive language model pre-training with keyphrase-guided infilling benefits low-resource keyphrase generation. \citet{kulkarni-etal-2022-learning} use keyphrase generation as a pre-training task to learn strong BART-based representations. 
\section{Conclusion}
This paper presents an empirical study of using PLMs for keyphrase extraction and generation. We investigate encoder-only vs. encoder-decoder PLMs, PLMs in different pre-training domains, and parameter allocation strategies for seq2seq PLMs. Our findings suggest that large or in-domain seq2seq PLMs approach the SOTA keyphrase generation performance, while with in-domain encoder-only PLMs, we can build strong and data-efficient keyphrase generation models. We demonstrate that the encoder has a more important role and a deep-encoder shallow-decoder approach empirically works well. Finally, we introduce strong in-domain PLMs and show their advantage in keyphrase generation. A comparison with KeyBART suggests that task-specific pre-training does not waive the need for in-domain pre-training. Future studies may further investigate PLMs with more parameters, keyphrase generation in other domains, and using SciBART, NewsBART, and NewsBERT in other downstream NLP applications.

\section*{Limitations and Ethical Statement}
Due to the constraints on computational power, we do not study large language models with more than one billion parameters. We hope future work can continue studying the effect of scaling up the model. In addition, our study only covers several domains for keyphrase generation. It is interesting to see whether our results can further generalize to more domains and languages. Finally, although we have thoroughly tested SciBART, NewsBART, and NewsBERT on keyphrase generation, we do not study them on other NLP tasks.

S2ORC and OAGKX are released under the Creative Commons By 4.0 License, and RealNews is released under Apache 2.0. We perform text cleaning and email and URL filtering on S2ORC and RealNews to remove sensitive information, and we keep OAGKX as-is. We use the SciKP and KPTimes benchmarking datasets distributed by the original authors. No additional preprocessing is performed before fine-tuning except lower-casing and tokenization. We do not re-distribute any of the pre-training and benchmark datasets. 

Potential risks of SciBART include accidental leakage of (1) sensitive personal information and (2) inaccurate factual information. For (1), we carefully preprocess the data in the preprocessing stage to remove personal information, including emails and URLs. However, we had difficulties desensitizing names and phone numbers in the text because they overlapped with the informative content. For (2), since SciBART is pre-trained on scientific papers, it may generate scientific-style statements that include inaccurate information. We encourage the potential users of SciBART not to rely fully on its outputs without verifying their correctness.

Pre-training SciBART, NewsBART, and NewsBERT are computationally heavy, and we estimate the total CO$_2$ emission to be around 600 kg using the \href{https://mlco2.github.io/impact/#compute}{calculation application} provided by \citet{1910.09700}. In addition, we estimate that all the fine-tuning experiments, including hyperparameter optimization, emitted around 1500 kg CO$_2$. We release the hyperparameters in the appendix section \ref{appendix-hyperparams} to help the community reduce the energy spent optimizing PLMs for various NLP applications in the science and the news domain.

\section*{Acknowledgment}
The research is partly supported by Taboola and an Amazon AWS credit award. We thank the Taboola team for helpful discussions and feedback. We also thank the anonymous ARR reviewers and the members of the UCLA-NLP group for providing their valuable feedback.

\bibliography{anthology,custom}
\bibliographystyle{acl_natbib}

\clearpage
\appendix

\twocolumn[{%
 \centering
 \Large\bf Supplementary Material: Appendices \\ [20pt]
}]

\section{Domain-specific BERT-like PLMs}

In Figure \ref{various-bert-models}, we list a range of publicly available in-domain BERT-like PLMs. We notice that previous studies do not consider utilize the knowledge from these PLMs to build better keyphrase generation models. 

\begin{figure}[h]
    \centering
    \small
    \resizebox{\linewidth}{!}{
    \begin{tabular}{ p{0.98\linewidth} }
    \toprule
    \textbf{Academic domain} \\
    \hdashline \\
    [-2ex]
    BioBERT \citep{Lee_2019}, ChemBERTa \citep{2010.09885}, [Bio|CS]\_RoBERTa \citep{gururangan-etal-2020-dont}, SciBERT \citep{beltagy-etal-2019-scibert}, PubMedBERT \citep{Gu_2022}, MatSciBERT \citep{2109.15290} \\
    \midrule \\ 
    [-2ex]
    \textbf{Social domain}  \\
    \hdashline \\
    [-2ex]
    ClinicalBERT \citep{alsentzer-etal-2019-publicly}, FinBERT \citep{2006.08097}, LEGAL-BERT \citep{chalkidis-etal-2020-legal}, JobBERT \citep{zhang-etal-2022-skillspan}, PrivBERT \citep{srinath-etal-2021-privacy}, SportsBERT \citep{sportsbert} \\
    \midrule \\ 
    [-2ex] 
    \textbf{Web domain}  \\
    \hdashline \\
    [-2ex]
    Twitter-roberta \citep{barbieri-etal-2020-tweeteval}, BERTweet \citep{nguyen-etal-2020-bertweet},  [News|Reviews]\_RoBERTa \citep{gururangan-etal-2020-dont}, HateBERT \citep{caselli-etal-2021-hatebert} \\
    \bottomrule
    \end{tabular}}
    \vspace{-2mm}
    \caption{
    Domain-specific encoder-only PLMs are available in a variety of domains. No prior work considered using these "domain experts" for keyphrase generation.
    }
    \label{various-bert-models}
\end{figure}


\begin{figure*}[t]
\centering
\includegraphics[width=0.98\textwidth]{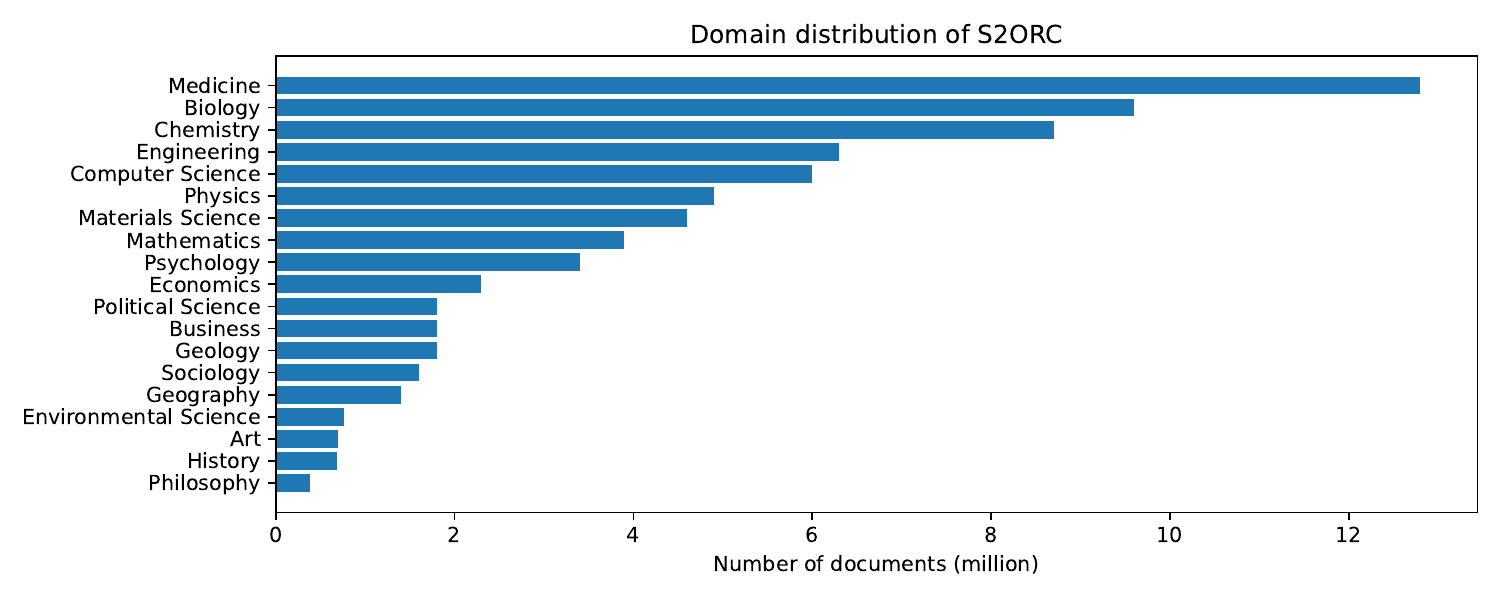}
\caption{Domain distribution of the S2ORC dataset.}
\label{s2orc-distribution}
\end{figure*}

\section{Pre-training Details}
\label{appendix_plm_pretraining}
\subsection{SciBART}
\paragraph{Corpus and Data Preprocessing} The S2ORC dataset contains over 100M papers from a variety of disciplines (Figure \ref{s2orc-distribution}). We train on all the titles and abstracts to increase the coverage of different topics. After removing non-English\footnote{We use \href{https://pypi.org/project/guess_language-spirit/}{guess\_language} for language detection.} or title-only entries, we fix wrong Unicode characters, remove emails and URLs, and convert the text to ASCII encoding\footnote{We use \href{https://github.com/jfilter/clean-text}{clean-text} for data cleaning.}. The final dataset contains 171.7M documents or 15.4B tokens in total. We reserve 10k documents for validation and 10k for testing and use the rest as training data.

\paragraph{Vocabulary} \citet{beltagy-etal-2019-scibert} suggest that using a domain-specific vocabulary is crucial to downstream in-domain fine-tuning performance. Following their approach, we build a cased BPE vocabulary in the scientific domain using the SentencePiece\footnote{  \url{https://github.com/google/sentencepiece}} library on the cleaned training data. We set the vocabulary size to 30K.

\paragraph{Training} For the pre-training objective, we only use text infilling as introduced in \citet{lewis-etal-2020-bart}. We mask 30\% of all tokens in each example, with the spans randomly sampled from a Poisson distribution ($\lambda=3.5$). For 10\% of the spans selected to mask, we replace them with a random token instead of the mask token. We set the maximum sequence length to 512. The model is pre-trained for 250k steps with batch size 2048, learning rate 3e-4, 10k warm-up steps, and polynomial learning rate decay. We use the Adam optimizer for pre-training. Using 8 Nvidia A100 GPUs, the training process took approximately eight days. 

\subsection{NewsBART}
The RealNews dataset contains over 32M documents and 19B tokens. Unlike news\_roberta \citep{gururangan-etal-2020-dont}, which is trained on an 11M document subset, we train on the entire dataset. We fix wrong Unicode characters, remove emails and URLs, and convert the text to ASCII encoding using the same tools we use for SciBART. Long documents are split into multiple chunks with no longer than 510 tokens. The model is pre-trained using the same setup as SciBART, the only difference being that NewsBART uses the pre-trained BART-base checkpoint. This is because BART-base is pre-trained with CC-News and already has substantial knowledge of the news domain. 

\subsection{NewsBERT}
We use the same preprocessed RealNews dataset to pre-train NewsBERT starting from BERT-base-uncased. Masked language modeling with 15\% dynamic masking is used as the objective. The model is pre-trained for 250k steps with batch size 512, learning rate 1e-4, 5k warm-up steps, and linear learning rate decay. We use the Adam optimizer for pre-training. Using 8 Nvidia V100 GPUs, the training process took approximately two days. 

\section{Dataset Statistics}
\label{appendix-data-stats}

Table \ref{tab:test-sets-statistics} summarizes the statistics of all testing datasets we use. In addition, we present the topic distribution of the S2ORC dataset in Figure \ref{s2orc-distribution}.

\setlength{\tabcolsep}{3pt}
\begin{table}[t]
    \centering
    {%
    \begin{tabular}{l | c  c  c c}
    \hline
    Dataset & \#Examples & \#KP & $\%$AKP & |KP|  \\
    \hline
    KP20k & 20000 & 5.3 & 37.1 & 2.0  \\
    Inspec & 500 & 9.8 & 26.4 & 2.5 \\
    Krapivin & 400 & 5.9 & 44.3 & 2.2\\
    NUS & 211 & 11.7 & 45.6 & 2.2  \\
    SemEval & 100 & 14.7 & 57.4 & 2.4 \\
    \hdashline
    KPTimes & 20000 & 5.0 & 37.8 & 2.0 \\
    \hdashline
    StackEx & 16000 & 2.7 & 42.5 & 1.3 \\
    \hline
    \end{tabular}
    }
    \caption{Test sets statistics. \#KP, $\%$AKP, and |KP| refers to the average number of keyphrases per document, the percentage of absent keyphrases, and the average number of words that each keyphrase contains.}
    \label{tab:test-sets-statistics}
\end{table}

\section{Baselines and Implementation}
\label{appendix_baselines}

\paragraph{Keyphrase Extraction} We further compare with a range of baselines including statistical methods \textbf{YAKE} \citep{Campos2018YAKECA} and \textbf{KP-Miner} \citep{el-beltagy-rafea-2010-kp}, graph-based unsupervised methods \textbf{TextRank} \citep{textrank2004}, \textbf{SingleRank} \citep{wan-xiao-2008-collabrank}, \textbf{PositionRank} \citep{florescu-caragea-2017-positionrank}, and \textbf{MultipartiteRank} \citep{boudin-2018-unsupervised}, as well as embedding-based unsupervised methods \textbf{EmbedRank} \citep{bennani-smires-etal-2018-simple}, \textbf{SIFRank+} \citep{8954611}, and the recent method \citet{liang-etal-2021-unsupervised} which combines BERT embedding and graph structure. We also compare with a supervised feature-based model \textbf{Kea} \citep{Kea1999}. We use \citet{boudin:2016:COLINGDEMO}'s public implementations for most of these baselines. For EmbedRank and SIFRank, we use the authors' public implementations. We implement our own version of \citet{liang-etal-2021-unsupervised}'s approach. We tune the hyperparameters of these methods using the KP20k and KPTimes validation set. 

\paragraph{Keyphrase Generation} For CatSeq, we run experiments using the publicly available implementation of \citet{chan-etal-2019-neural}.\footnote{\url{https://github.com/kenchan0226/keyphrase-generation-rl}} For ExHiRD, Transformer, and SetTrans, we use the authors' implementations to measure the performance. We use the earliest version of KeyBART available at \url{https://zenodo.org/record/5784384#.Y0eToNLMJcA}.

\section{Implementation details of PLM-base keyphrase generation and extraction}
\label{appendix_plm_implementations}
\paragraph{Keyphrase Extraction} We implemented our models with Huggingface Transformers\footnote{\url{https://github.com/huggingface/transformers}} and TorchCRF\footnote{\url{https://github.com/s14t284/TorchCRF}}. The models are trained for ten epochs with early stopping. We use a learning rate of 1e-5 with linear decay and batch size 32 for most models (see appendix for all the hyperparameters). We use AdamW with $\beta_1=0.9$ and $\beta_2=0.999$.

\paragraph{Keyphrase Generation}
For BART and T5, we use Huggingface Transformers and train for 15 epochs with early stopping. We use learning rate 6e-5, polynomial decay, batch size 64, and the AdamW optimizer. To fine-tune SciBART-base, SciBART-large, and NewsBART-base, we use the Translation task provided by fairseq\footnote{\url{https://github.com/facebookresearch/fairseq}} and train for 10 epochs. We use learning rate 3e-5, polynomial decay, and the AdamW optimizer. 

For BERT-G, SciBERT-G, NewsBART-G, RoBERTa-G, and UniLM, we base on \citet{arxiv.1905.03197}'s implementations \footnote{\url{https://github.com/microsoft/unilm}}. For most models, we train for 20k steps with batch size 128, learning rate 1e-4, and linear decay. We set the maximum source and target length to 464 tokens and 48 tokens, respectively. We mask 80\% of the target tokens and randomly replace an additional 10\%. We use the AdamW optimizer.

We use greedy decoding for all the models. The fine-tuning experiments are run on a local GPU server with Nvidia GTX 1080 Ti and RTX 2080 Ti GPUs. We use at most 4 GPUs and gradient accumulation to achieve the desired batch sizes.

\section{Hyperparameter Optimization}
\label{appendix-hyperparams}
For each of the PLM-based keyphrase extraction and keyphrase generation methods, we perform a careful hyperparameter search over the learning rate, learning rate schedule, batch size, and warm-up steps. The corresponding search spaces are \{1e-5, 5e-4\}, \{linear, polynomial\}, \{16, 32, 64, 128\}, and \{500, 1000, 2000, 4000\}. The best hyperparameters found are presented in Table \ref{tab:hyperparams}. 

\begin{table*}[h]
    \centering
    \begin{tabular}{l | c | c | c | c | c | c | c | c }
    \hline
    Model & dropout & wdecay & optimizer & bsz & \#epoch & \#warm-up & lr & lr schedule \\
    \hline
    \multicolumn{6}{l}{\textbf{Keyphrase extraction}} \\
    \hdashline
    Transformer & 0.1 & 0.01 & AdamW & 32 & 10 & 2000 & 3e-5 & linear \\
    BERT  & 0.1 & 0.01 & AdamW & 32 & 10 & 1000 & 1e-5 & linear \\
    SciBERT  & 0.1 & 0.01 & AdamW & 32 & 10 & 1000 & 1e-5 & linear \\
    RoBERTa & 0.1 & 0.01 & AdamW & 32 & 10 & 1000 & 1e-5 & linear \\
    Transformer+CRF & 0.1 & 0.01 & AdamW &32 & 10 & 2000 & 3e-5 & linear \\
    BERT+CRF  & 0.1 & 0.01 & AdamW & 32 & 10 & 2000 & 1e-5 & linear \\
    SciBERT+CRF  & 0.1 & 0.01 & AdamW & 32 & 10 & 2000 & 1e-5 & linear \\
    RoBERTa+CRF & 0.1 & 0.01 & AdamW & 32 & 10 & 2000 & 1e-5 & linear \\
    \hline
    \multicolumn{6}{l}{\textbf{Keyphrase generation}} \\
    \hdashline
    BERT-G & 0.1 & 0.01 & AdamW & 64 & 6 & 4000 & 1e-4 & linear\\
    SciBERT-G & 0.1 & 0.01 & AdamW & 128 & 6 & 2000 & 1e-4 & linear \\
    RoBERTa-G & 0.1 & 0.01 & AdamW & 64 & 6 & 4000 & 1e-4 & linear\\
    UniLM & 0.1 & 0.01 & AdamW & 128 & 6 & 2000 & 1e-4 & linear \\
    BERT2BERT & 0.0 & 0.01 & AdamW & 32 & 20 & 2000 & 5e-5 & linear \\
    BART-base & 0.1 & 0.01 & AdamW & 64 & 15 & 2000 & 6e-5 & polynomial \\
    SciBART-base & 0.1 & 0.01 & AdamW & 32 & 10 & 2000 & 3e-5 & polynomial \\
    T5-base & 0.1 & 0.01 & AdamW & 64 & 15 & 2000 & 6e-5 & polynomial \\
    KeyBART & 0.1 & 0.01 & AdamW & 64 & 15 & 2000 & 3e-5 & polynomial \\
    \hline
    \end{tabular}
    \caption{Hyperparameters for fine-tuning PLMs for keyphrase extraction and keyphrase generation on KP20k. The hyperparameters are determined using the loss on the KP20k validation dataset. We follow a similar set of hyperparameters for KPTimes. "wdecay" = weight decay, "bsz" = batch size, "\#warm-up" = the number of warm-up steps, "lr" = learning rate, "lr schedule" = learning rate decay schedule. We use early stopping for all the models and use the model with the lowest validation loss as the final model. }
    \label{tab:hyperparams}
\end{table*}

\section{Data Efficiency on KPTimes}
\begin{figure}[t!]
\centering
\vspace{-6mm}
\includegraphics[width=0.48\textwidth]{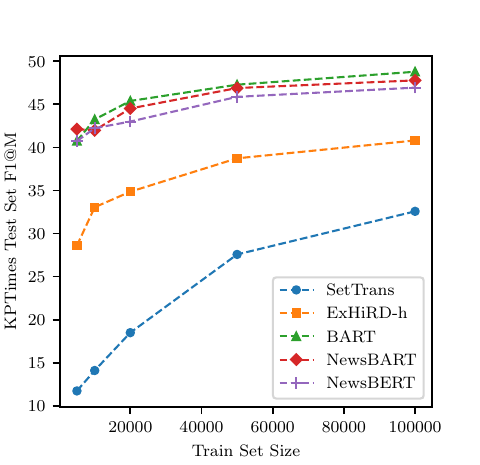}
\vspace{-6mm}
\caption{Present keyphrase generation performance as a function of training set size.}
\vspace{-2mm}
\label{perf-vs-resource-kptimes}
\end{figure}

Similar to the experiment presented in Figure \ref{perf-vs-resource-kp20k}, we conduct an experiment of training different models on low-resource subsets of KPTimes. The results are presented in Figure \ref{perf-vs-resource-kptimes}. In general, we find a similar pattern: PLMs can greatly outperform methods trained from scratch in low-resource settings. Note that BART should be considered partially in-domain because it is pre-trained on CC-NEWS. 

\section{Prefix for T5 models}
We experimented with using different prefixes for using T5 models for seq2seq keyphrase generation: (1) ``summarize: ''; (2) ``generate keyphrases: ''; and (3) an empty string (None).  Our results are summarized in Table \ref{tab:t5-prefixes}.  We empirically find that (1) and (2) both produce the best results, while (3) gives a worse performance. The result is intuitive because T5 models have been pre-trained on different tasks specified in the form of prefixes.
\setlength{\tabcolsep}{3.5pt}
\begin{table}[h]
    \centering
    \resizebox{\linewidth}{!}{%
    \begin{tabular}{l | c c | c c  }
    \hline
    \multirow{2}{*}{Prefix} & \multicolumn{2}{c|}{\textbf{Present KPs}} & \multicolumn{2}{c}{\textbf{Absent KPs}} \\
    & F1@5 & F1@M & F1@5 & F1@M \\
    \hline
    \texttt{summarize:} & 33.6 & 38.8 & \textbf{1.7} & \textbf{3.4} \\
    \hdashline
    \begin{tabular}[x]{@{}l}\texttt{generate}\\\texttt{keyphrases:}\end{tabular} & \textbf{33.7} & \textbf{38.9} & \textbf{1.7} & \textbf{3.4} \\
    \hdashline
    \texttt{None} & 33.5 & 37.9 & 1.4 & 2.9 \\
    \hline
    \end{tabular}
    }
    \caption{
    Comparison across different prefixes for fine-tuning T5. We use T5-base for all experiments and report the averaged scores across three runs. 
    }
    \label{tab:t5-prefixes}
\end{table}

\section{Encoder Quality of B2B models}
\label{appendix-encoder-quality}
\setlength{\tabcolsep}{3.5pt}
\begin{table}[!t]
    \small
    \resizebox{\linewidth}{!}{%
    \begin{tabular}{l | c | c c | c c  }
    \hline
    \multirow{2}{*}{Model Size} & \multirow{2}{*}{|M|} & \multicolumn{2}{c|}{\textbf{freeze+CRF}} & \multicolumn{2}{c}{\textbf{unfreeze}} \\
    & & F1@5 & F1@M & F1@5 & F1@M \\
    \hline
    2 layers & 39M & 19.2 & 30.5 & 31.5 & 45.2 \\
    4 layers & 53M & 26.5 & 38.1 & 32.6 & 46.7 \\
    6 layers & 67M & \textbf{27.4} & \textbf{39.1} & 33.2 & 47.8 \\
    8 layers & 81M & 26.8 & 38.7 & \textbf{34.4} & \textbf{48.5} \\
    10 layers & 95M & 26.9 & 38.3 & 33.0 & 47.9 \\
    \hline
    \end{tabular}
    }
    \caption{
    The feature quality of the encoders via sequence labeling results on the KPTimes test set. The models are the encoders taken from BERT2BERT models trained on keyphrase generation and further trained on keyphrase extraction on KPTimes. "freeze" means freezing the underlying encoder model while "unfreeze" means fine-tuning the entire model. For the unfreeze version we found using CRF unnecessary.
    }
    \label{tab:feature_quality_b2b_enc}
\end{table}
To further investigate the nature of the encoder representation after being trained in a BERT2BERT formulation on keyphrase generation, we separate the encoder's weights and use it as a feature extractor. Concretely, we fix the encoder weights, add a CRF layer on top of it, and train on keyphrase extraction via sequence labeling on KPTimes for 5 epochs. The results are summarized in Table \ref{tab:feature_quality_b2b_enc}. We find that the encoders of BERT2BERT keyphrase generation models indeed build a strong representation such that simply fine-tuning a linear classifier on the top can achieve non-trivial keyphrase extraction performance. Furthermore, the encoder's quality is positively related to the corresponding BERT2BERT performance.

\section{Inference Speed}
\label{appendix-inference-speed}
\begin{figure}[t]
\centering
\vspace{-10pt}
\includegraphics[width=0.48\textwidth]{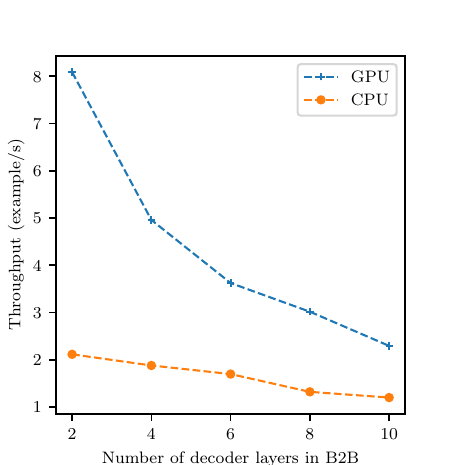}
\caption{Inference speed of BERT2BERT models with different encoder-decoder configurations on GPU and CPU. All models have 12 layers in total. The model with x decoder layers has 12-x encoder layers.}
\label{b2b-inference-speed}
\end{figure}

To quantify the inference speed of different BERT2BERT configurations, we measure and compare the inference throughput of B2B-2+10, B2B-4+8, B2B-6+6, B2B-8+4, and B2B-10+2 on GPU and CPU. We use the best model trained on KP20k and test on the KP20k test set with batch size 1, no padding, and no speedup libraries. We use an Nvidia GTX 1080 Ti card for GPU and test on the full KP20k test set. We use a local server with 40 cores for CPU and test on the first 1000 examples from the KP20k test set. We report the averaged throughput (in example/s) across three runs in Figure \ref{b2b-inference-speed}. We observe that the throughput decreases significantly with deeper decoders for both CPU and GPU. Our recommended B2B-8+4 configuration achieves better performance than 6+6 while being 37\% faster on GPU and 11\% faster on CPU.

\section{Artifact Release}
To facilitate future research, we plan to release our pre-trained SciBART, NewsBART, and NewsBERT checkpoints as well as the SciBART fine-tuned on OAGK. We will limit access to the models with a non-commercial license. We will also release the raw predictions of our models to enable fair comparisons in future work. 

\section{All Experiment Results}
We summarize all of our experiment results on SciKP and KPTimes in Table \ref{tab:scikp-all-results-pkp}, Table \ref{tab:scikp-all-results-akp}, Table \ref{tab:other-all-results-pkp}, and Table \ref{tab:other-all-results-akp}.

\setlength{\tabcolsep}{4pt}
\begin{table*}[!ht]
    \centering
    \resizebox{\linewidth}{!}{%
    \begin{tabular}{l | c | l l | l l | l l | l l | l l }
    \hline
    \multirow{2}{*}{Method} &
    \multirow{2}{*}{|M|} & \multicolumn{2}{c|}{\textbf{KP20k}} & \multicolumn{2}{c|}{\textbf{Inspec}} & \multicolumn{2}{c|}{\textbf{Krapivin}} & \multicolumn{2}{c|}{\textbf{NUS}} & \multicolumn{2}{c}{\textbf{SemEval}} \\
    & & F1@5 & F1@M & F1@5 & F1@M & F1@5 & F1@M & F1@5 & F1@M & F1@5 & F1@M\\
    \hline
    \multicolumn{8}{l}{(keyphrase extraction baselines)} \\
    KP-Miner & - & 21.9 & \hfil - \hfil & 14.4 & \hfil - \hfil & 20.6 & \hfil - \hfil& 28.3 & \hfil - \hfil & 24.4 & \hfil - \hfil\\
    YAKE & - & 18.8 & \hfil - \hfil & 19.0 & \hfil - \hfil & 19.4 & \hfil - \hfil & 22.4 & \hfil - \hfil & 20.7 & \hfil - \hfil\\
    TextRank & - & 16.2 & \hfil - \hfil & 22.6 & \hfil - \hfil & 13.6 & \hfil - \hfil& 20.8 & \hfil - \hfil & 18.7 & \hfil - \hfil \\
    PositionRank & - & 18.9 & \hfil - \hfil& 30.4 & \hfil - \hfil & 18.9 & \hfil - \hfil& 23.0 & \hfil - \hfil& 23.8 & \hfil - \hfil\\
    MultipartiteRank & - & 18.8 & \hfil - \hfil& 25.9 & \hfil - \hfil & 17.4 & \hfil - \hfil& 24.8 & \hfil - \hfil& 22.2 & \hfil - \hfil\\
    EmbedRank & - & 15.5 & \hfil - \hfil& 33.6 & \hfil - \hfil& 16.9 & \hfil - \hfil& 17.3 & \hfil - \hfil& 19.2 & \hfil - \hfil\\
    SIFRank+ & - & 20.0 & \hfil - \hfil & 35.1 & \hfil - \hfil& 19.6 & \hfil - \hfil& 25.5 & \hfil - \hfil& 24.8 & \hfil - \hfil\\
    \citet{liang-etal-2021-unsupervised} & - & 17.7 & \hfil - \hfil& 29.6 & \hfil - \hfil& 16.9 & \hfil - \hfil& 25.0 & \hfil - \hfil& 25.3 & \hfil - \hfil\\
    Kea & - & 19.4 &\hfil - \hfil& 12.7 & \hfil - \hfil& 16.0 & \hfil - \hfil& 23.6 & \hfil - \hfil& 15.5 & \hfil - \hfil\\
    \hline
    \multicolumn{6}{l}{(supervised keyphrase extraction)}  \\
    Transformer & 110M & 23.5$_6$ & 33.8$_4$ & 11.1$_3$ & 15.4$_5$ & 16.6$_{11}$ & 26.5$_{16}$ & 26.1$_{13}$ & 35.9$_{13}$ & 18.7$_7$ & 25.2$_7$ \\
    Transformer+CRF & 110M & 24.9$_{10}$ & 36.4$_6$ & 13.3$_{12}$ & 18.7$_{13}$ & 18.9$_{19}$ & 29.7$_{22}$ & 27.8$_{15}$ & 37.7$_{16}$ & 19.8$_{20}$ & 27.3$_{20}$ \\
    BERT-base & 110M & 27.9$_1$ & 38.9$_1$ & 12.8$_4$ & 17.4$_4$ & 20.7$_{11}$ & 30.2$_{17}$ & 30.9$_7$ & 41.0$_6$ & 21.8$_{11}$ & 28.5$_{13}$ \\
    BERT-base+CRF & 110M & 28.0$_2$ & 40.6$_3$ & 13.7$_5$ & 18.8$_9$ & 21.0$_3$ & 32.6$_{10}$ & 31.3$_9$ & 41.9$_{16}$ & 22.3$_6$ & 29.3$_{15}$  \\
    SciBERT & 110M & 28.6$_9$ & 40.5$_5$ & 13.1$_{11}$ & 17.8$_{13}$ & 19.9$_{12}$ & 30.3$_{13}$ & 29.7$_5$ & 39.0$_{13}$ & 20.0$_{22}$ & 26.3$_{30}$ \\
    SciBERT+CRF & 110M & 28.4$_{13}$ & 42.1$_{11}$ & 13.9$_{14}$ & 19.6$_{11}$ & 20.6$_{12}$ & 32.2$_{14}$ & 29.9$_{12}$ & 40.8$_{10}$ & 21.3$_{15}$ & 28.6$_{12}$  \\
    NewsBERT & 110M & 25.8$_6$ & 37.5$_5$ & 11.8$_4$ & 16.4$_3$ & 18.3$_6$ & 28.1$_{10}$ & 27.5$_{11}$ & 37.6$_{13}$ & 19.7$_2$ & 26.4$_7$ \\
    NewsBERT+CRF & 110M & 26.8$_3$ & 39.7$_1$ & 13.5$_4$ & 19.0$_8$ & 19.9$_{11}$ & 31.4$_{19}$ & 29.5$_{12}$ & 40.4$_{13}$ & 21.5$_5$ & 29.0$_6$ \\
    RoBERTa-base & 125M & 27.9$_4$ & 39.4$_2$ & 13.9$_8$ & 18.9$_{12}$ & 19.1$_3$ & 29.5$_9$ & 29.6$_{15}$ & 38.7$_{17}$ & 20.7$_{18}$ & 25.8$_{13}$ \\
    RoBERTa-base+CRF & 125M & 26.7$_6$ & 39.0$_5$ & 12.5$_4$ & 17.5$_5$ & 18.7$_3$ & 29.3$_1$ & 28.7$_6$ & 39.5$_7$ & 20.1$_{10}$ & 26.8$_{11}$ \\
    \hline
    \multicolumn{8}{l}{(supervised keyphrase generation)} \\
    CatSeq & 21M & 29.1 & 36.7 & 22.5 & 26.2 & 26.9 & 35.4 & 32.3 & 39.7 & 24.2 & 28.3 \\
    ExHiRD-h & 22M & 31.1$_{1}$ & 37.4$_{0}$ & 25.4$_{4}$ & 29.1$_{3}$ & 28.6$_{4}$ & 30.8$_{4}$ & \hfil - \hfil & \hfil - \hfil & 30.4$_{17}$ & 28.2$_{18}$ \\
    Transformer & 98M & 33.3$_{1}$ & 37.6$_{2}$ & 28.8$_{7}$ & 33.3$_{5}$ & 31.4$_{9}$ & 36.5$_{7}$ & 37.8$_{6}$ & 42.9$_{9}$ & 28.8$_{5}$ & 32.1$_{8}$ \\
    SetTrans & 98M & 35.6$_{0}$ & 39.1$_{2}$ & 29.1$_{3}$ & 32.8$_{1}$ & 33.5$_{10}$ & 37.5$_{11}$ & 39.9$_{8}$ & 44.6$_{22}$ & 32.2$_{8}$ & 34.2$_{14}$ \\
    \hdashline
    BERT-G & 110M & 31.3$_{6}$ & 37.9$_{2}$ & 25.9$_{5}$ & 31.3$_{5}$ & 26.3$_{5}$ & 32.2$_{2}$ & 35.2$_{9}$ & 40.9$_{8}$ & 26.3$_{11}$ & 31.0$_{12}$ \\
    RoBERTa-G & 125M & 28.8$_{5}$ & 36.9$_{3}$ & 22.0$_{5}$ & 27.4$_{7}$ & 23.5$_{6}$ & 31.3$_{12}$ & 30.9$_4$ & 38.1$_{7}$ & 23.3$_{3}$ & 28.6$_{9}$ \\
    SciBERT-G  & 110M & 32.8$_{1}$ & 39.7$_{1}$ & 25.7$_{1}$ & 31.3$_{4}$ & 27.2$_{3}$ & 33.4$_{6}$ & 35.8$_{18}$ & 41.5$_{17}$ & 24.7$_{2}$ & 28.4$_{7}$ \\
    NewsBERT-G & 110M & 29.9$_{5}$ & 36.8$_{2}$ & 26.9$_{7}$ & 32.5$_{8}$ & 25.9$_{6}$ & 31.0$_{9}$ & 33.7$_{9}$ & 39.5$_{10}$ & 24.9$_{13}$ & 29.6$_{15}$ \\    
    UniLM & 110M & 26.7$_{6}$ & 34.6$_{3}$ & 18.2$_{17}$ & 23.6$_{24}$ & 23.5$_{6}$ & 28.5$_{21}$ & 28.4$_{6}$ & 35.3$_{5}$ & 21.5$_{10}$ & 26.8$_{17}$ \\
    B2B-2+10 & 158M & 30.4$_{1}$ & 36.4$_{1}$ & 26.0$_{9}$ & 31.3$_{11}$ & 27.6$_{2}$ & 33.1$_{3}$ & 36.0$_{5}$ & 41.0$_{7}$ & 27.4$_{3}$ & 31.1$_{7}$ \\
    B2B-4+8 & 153M & 31.7$_{1}$ & 37.7$_{2}$ & 26.5$_{5}$ & 31.7$_{6}$ & 27.1$_{10}$ & 32.5$_{5}$ & 35.6$_{3}$ & 40.3$_{6}$ & 26.0$_{16}$ & 30.5$_{17}$ \\
    B2B-6+6 & 148M & 32.1$_{2}$ & 37.7$_{1}$ & 26.7$_{8}$ & 31.7$_{7}$ & 27.3$_{7}$ & 31.6$_{7}$ & 35.4$_{4}$ & 40.3$_{7}$ & 26.4$_{3}$ & 29.8$_{12}$ \\
    B2B-8+4 & 143M & 32.2$_{2}$ & 38.0$_{0}$  & 26.0$_{1}$ & 30.9$_{0}$ & 27.2$_{4}$ & 32.1$_{6}$ & 36.4$_{15}$ & 41.8$_{12}$ & 28.0$_{11}$ & 32.8$_{9}$ \\
    B2B-10+2 & 139M & 31.7$_{2}$ & 38.0$_{2}$  & 26.4$_{8}$ & 31.8$_{11}$ & 26.4$_{6}$ & 31.3$_{8}$ & 34.4$_{22}$ & 39.4$_{16}$ & 26.0$_{20}$ & 30.0$_{15}$\\
    \hdashline
    BART-base & 140M & 32.2$_2$ & 38.8$_3$ & 27.0$_3$ & 32.3$_7$ & 27.0$_6$ & 33.6$_6$ & 36.6$_1$ & 42.4$_8$ & 27.1$_{11}$ & 32.1$_{21}$ \\
    BART-large & 406M & 33.2$_{4}$ & 39.2$_{2}$ & 27.6$_{11}$ & 33.3$_{9}$ & 28.4$_{2}$ & 34.7$_{3}$ & 38.0$_{8}$ & 43.5$_{11}$ & 27.4$_{12}$ & 31.1$_{16}$ \\
    T5-base & 223M & 33.6$_{1}$ & 38.8$_{0}$ & 28.8$_{5}$ & 33.9$_{5}$ & 30.2$_{3}$ & 35.0$_{2}$ & 38.8$_{6}$ & 44.0$_{4}$ & 29.5$_{16}$ & 32.6$_{16}$ \\
    T5-large & 770M & 34.3$_{2}$ & 39.3$_{0}$ & 29.5$_{1}$ & 34.3$_{4}$ & 31.5$_{2}$ & 35.9$_{5}$ & 39.8$_{4}$ & 43.8$_{6}$ & 29.7$_{10}$ & 32.1$_{11}$ \\ 
    KeyBART & 406M & 32.5$_{1}$ & 39.8$_{2}$ & 26.8$_{3}$ & 32.5$_{5}$ & 28.7$_{6}$ & 36.5$_{14}$ & 37.3$_{7}$ & 43.0$_{10}$ & 26.0$_{8}$ & 28.9$_{4}$ \\
    SciBART-base & 124M & 34.1$_{1}$ & 39.6$_{2}$ & 27.5$_{10}$ & 32.8$_{8}$ & 28.2$_{8}$ & 32.9$_{11}$ & 37.3$_{7}$ & 42.1$_{14}$ & 27.0$_{8}$ & 30.4$_{8}$ \\
    SciBART-base+OAGKX & 124M & 35.3$_{3}$ & 41.5$_{2}$ & 27.1$_{7}$ & 33.0$_{6}$ & 27.7$_{7}$ & 33.7$_{9}$ & 38.2$_{7}$ & 42.4$_{6}$ & 29.2$_{6}$ & 32.9$_{9}$ \\
    SciBART-large & 386M & 34.7$_{3}$ & 41.5$_{4}$ & 26.1$_{12}$ & 31.7$_{13}$ & 27.1$_{11}$ & 32.4$_{12}$ & 36.4$_{18}$ & 40.9$_{12}$ & 27.9$_{14}$ & 32.0$_{12}$ \\
    SciBART-large+OAGKX & 386M & 36.2$_{1}$ & 43.2$_{0}$ & 26.7$_{5}$ & 33.1$_{4}$ & 28.9$_{8}$ & 34.7$_{6}$ & 38.7$_{11}$ & 44.2$_{11}$ & 30.0$_{7}$ & 33.3$_{19}$ \\
    NewsBART-base & 140M & 32.4$_{3}$ & 38.7$_{2}$ & 26.2$_{10}$ & 31.7$_{11}$ & 26.2$_{8}$ & 32.3$_{15}$ & 36.9$_{8}$ & 42.4$_{10}$ & 26.4$_{21}$ & 30.4$_{23}$ \\
    \hline
    \end{tabular}
    }
    \caption{Present keyphrase evaluation results of all the methods on the SciKP benchmark. The reported results are averaged across three runs with different random seeds. The standard deviation of each entry is presented in the subscript. For example, 23.5$_6$ means an average of 23.5 with a standard deviation of 0.6. We omit the subscript for deterministic methods or methods with a single run.}
    \label{tab:scikp-all-results-pkp}
\end{table*}

\setlength{\tabcolsep}{4pt}
\begin{table*}[!ht]
    \centering
    \resizebox{\linewidth}{!} {%
    \begin{tabular}{l | c | l l | l l | l l | l l | l l }
    \hline
    \multirow{2}{*}{Method} &
    \multirow{2}{*}{|M|} & \multicolumn{2}{c|}{\textbf{KP20k}} & \multicolumn{2}{c|}{\textbf{Inspec}} & \multicolumn{2}{c|}{\textbf{Krapivin}} & \multicolumn{2}{c|}{\textbf{NUS}} & \multicolumn{2}{c}{\textbf{SemEval}} \\
    & & F1@5 & F1@M & F1@5 & F1@M & F1@5 & F1@M & F1@5 & F1@M & F1@5 & F1@M\\
    \hline
    CatSeq & 21M & 1.5 & 3.2 & 0.4 & 0.8 & 1.8 & 3.6 & 1.6 & 2.8 & 2.0 & 2.8 \\
    ExHiRD-h & 22M & 1.6$_{0}$ & 2.5$_{0}$ & 1.1$_{1}$ & 1.6$_{2}$ & 2.2$_{3}$ & 3.3$_{4}$ & \hfil - \hfil & \hfil - \hfil & 1.6$_{4}$ & 2.1$_{6}$ \\
    Transformer & 98M & 2.2$_{2}$ & 4.6$_{4}$ & 1.2$_{0}$ & 2.3$_{1}$ & 3.3$_{2}$ & 6.3$_{4}$ & 2.5$_{4}$ & 4.4$_{9}$ & 1.6$_{2}$ & 2.2$_{4}$ \\
    SetTrans & 98M & 3.5$_{1}$ & 5.8$_{1}$ & 1.9$_{1}$ & 3.0$_{1}$ & 4.5$_{1}$ & 7.2$_{3}$ & 3.7$_{10}$ & 5.5$_{17}$ & 2.2$_{2}$ & 2.9$_{2}$ \\
    \hdashline
    BERT-G & 110M  & 1.9$_{1}$ & 3.7$_{2}$ & 1.0$_{2}$ & 1.9$_{6}$ & 2.4$_{2}$ & 4.3$_{4}$ & 2.2$_{5}$ & 3.9$_{11}$ & 1.4$_{2}$ & 2.0$_{3}$ \\
    RoBERTa-G & 125M & 2.0$_{0}$ & 3.1$_{0}$ & 1.0$_{1}$ & 2.0$_2$ & 2.7$_2$ & 4.8$_3$ & 2.5$_4$ & 4.3$_8$ & 2.1$_1$ & 2.9$_1$ \\
    SciBERT-G & 110M & 2.4$_{0}$ & 4.6$_{1}$ & 1.4$_{2}$ & 2.7$_{5}$ & 2.4$_{3}$ & 4.6$_{5}$ & 3.4$_{9}$ & 5.9$_{18}$& 1.3$_{1}$ & 1.8$_{2}$ \\
    NewsBERT-G & 110M & 1.3$_{1}$ & 2.6$_{3}$ & 0.8$_{2}$ & 1.5$_{4}$ & 1.7$_{2}$ & 3.4$_{2}$ & 1.5$_{2}$ & 2.8$_{6}$ & 1.3$_{2}$ & 1.9$_{3}$ \\
    UniLM & 110M & 1.4$_{2}$ & 2.8$_{4}$ & 0.5$_{1}$ & 0.8$_{2}$ & 1.4$_{3}$ & 2.4$_{5}$ & 1.7$_{3}$ & 3.2$_{7}$ & 1.0$_{4}$ & 1.5$_{6}$ \\
    B2B-2+10 & 158M & 2.1$_{1}$ & 3.9$_{1}$ & 1.1$_{2}$ & 1.9$_{3}$ & 2.7$_{4}$ & 4.7$_{6}$ & 2.8$_{5}$ & 4.7$_{7}$ & 1.9$_{1}$ & 2.6$_{2}$ \\
    B2B-4+8 & 153M & 2.2$_{1}$ & 4.1$_{1}$ & 1.1$_{1}$ & 2.0$_{1}$ & 2.6$_{1}$ & 4.4$_{1}$ & 2.7$_{5}$ & 4.3$_{6}$ & 2.2$_{2}$ & 2.9$_{3}$ \\
    B2B-6+6 & 148M & 2.2$_{2}$ & 4.1$_{2}$ & 1.0$_{3}$ & 1.8$_{5}$ & 2.7$_{1}$ & 4.6$_{1}$ & 2.8$_{5}$ & 4.2$_{5}$ & 1.7$_{8}$ & 2.3$_{9}$ \\
    B2B-8+4 & 143M & 2.2$_{1}$ & 4.2$_{1}$ & 1.1$_{0}$ & 2.0$_{1}$ & 2.8$_{3}$ & 5.2$_{5}$ & 2.6$_{8}$ & 4.1$_{15}$ & 1.8$_{3}$ & 2.3$_{3}$ \\
    B2B-10+2 & 139M & 2.1$_{1}$ & 4.1$_{2}$ & 1.2$_{4}$ & 2.3$_{8}$ & 2.4$_{3}$ & 4.4$_{4}$ & 2.6$_{6}$ & 4.6$_{13}$ & 1.8$_{4}$ & 2.5$_{5}$ \\
    \hdashline
    BART-base & 140M & 2.2$_1$ & 4.2$_2$ & 1.0$_1$ & 1.7$_2$ & 2.8$_3$ & 4.9$_6$ & 2.6$_4$ & 4.2$_9$ & 1.6$_1$ & 2.1$_2$ \\
    BART-large & 406M & 2.7$_{2}$ & 4.7$_{2}$ & 1.5$_{3}$ & 2.4$_{4}$ & 3.1$_{1}$ & 5.1$_{2}$ & 3.1$_{5}$ & 4.8$_{9}$ & 1.9$_{3}$ & 2.4$_{3}$ \\
    T5-base & 223M & 1.7$_{0}$ & 3.4$_{0}$ & 1.1$_{1}$ & 2.0$_{3}$ & 2.3$_{2}$ & 4.3$_{4}$ & 2.7$_{0}$ & 5.1$_{3}$ & 1.4$_{4}$ & 2.0$_{5}$ \\
    T5-large & 770M & 1.7$_{0}$ & 3.5$_{0}$ & 1.1$_{3}$ & 2.1$_{6}$ & 2.3$_{4}$ & 4.5$_{7}$ & 2.5$_{3}$ & 4.2$_{6}$ & 1.5$_{1}$ & 2.0$_{3}$ \\ 
    KeyBART & 406M & 2.6$_{1}$ & 4.7$_{1}$ & 1.4$_{2}$ & 2.3$_{2}$ & 3.6$_{2}$ & 6.4$_{6}$ & 3.1$_{4}$ & 5.5$_{7}$ & 1.6$_{4}$ & 2.2$_{5}$ \\
    SciBART-base & 124M & 2.9$_{3}$ & 5.2$_{4}$ & 1.6$_{2}$ & 2.8$_{4}$ & 3.3$_{4}$ & 5.4$_{8}$ & 3.3$_{1}$ & 5.3$_{2}$ & 1.8$_{1}$ & 2.2$_{1}$ \\
    SciBART-base+OAGKX & 124M & 2.8$_{1}$ & 5.2$_{1}$ & 1.5$_{3}$ & 2.7$_{4}$ & 3.2$_{4}$ & 5.7$_{7}$ & 2.8$_{1}$ & 4.8$_{2}$ & 1.8$_{0}$ & 2.4$_{0}$ \\
    SciBART-large & 386M & 3.1$_{2}$ & 5.7$_{3}$ & 1.5$_{2}$ & 2.6$_{2}$ & 3.4$_{1}$ & 5.6$_{3}$ & 3.2$_{5}$ & 5.0$_{7}$ & 2.6$_{6}$ & 3.3$_{8}$ \\
    SciBART-large+OAGKX & 386M & 3.2$_{1}$ & 6.2$_{1}$ & 1.7$_{1}$ & 3.0$_{1}$ & 3.6$_{2}$ & 6.4$_{7}$ & 3.3$_{2}$ & 5.5$_{5}$ & 2.3$_{1}$ & 3.1$_{2}$ \\
    NewsBART-base & 140M & 2.2$_{1}$ & 4.4$_{2}$ & 1.0$_{1}$ & 1.8$_{2}$ & 2.4$_{2}$ & 4.5$_{4}$ & 2.4$_{4}$ & 4.0$_{9}$ & 1.6$_{1}$ & 2.2$_{2}$ \\
    \hline
    \end{tabular}
    }
    \caption{Absent keyphrase evaluation results of all keyphrase generation methods on the SciKP benchmark. The standard deviation of each entry is presented in the subscript. For example, 23.5$_6$ means an average of 23.5 with a standard deviation of 0.6. We omit the subscript for deterministic methods or methods with a single run.}
    \label{tab:scikp-all-results-akp}
\end{table*}

\setlength{\tabcolsep}{4pt}
\begin{table*}[]
    \centering
    \begin{tabular}{l | c | l l | l l }
    \hline
    \multirow{2}{*}{Method} &
    \multirow{2}{*}{|M|} & \multicolumn{2}{c|}{\textbf{KPTimes}} & \multicolumn{2}{c}{\textbf{StackEx}} \\
     & & F1@5 & F1@M & F1@5 & F1@M \\
    \hline
    \multicolumn{4}{l}{(keyphrase extraction baselines)} \\
    KP-Miner & - & 18.0 & \hfil - \hfil & 16.8 & \hfil - \hfil \\
    YAKE & - & 13.1 & \hfil - \hfil & 13.0 & \hfil - \hfil \\
    TextRank & - & 17.4 & \hfil - \hfil & 12.6 & \hfil - \hfil \\
    PositionRank & - & 11.9 & \hfil - \hfil & 12.1 & \hfil - \hfil \\
    MultipartiteRank & - & 19.5 & \hfil - \hfil & 13.7 & \hfil - \hfil \\
    EmbedRank & - & 10.2 & \hfil - \hfil & 11.4 & \hfil - \hfil \\
    SIFRank+ & - & 15.8 & \hfil - \hfil & 12.0  & \hfil - \hfil \\
    \citet{liang-etal-2021-unsupervised} & - & 16.2 & \hfil - \hfil & 13.8 & \hfil - \hfil \\
    Kea & - & 18.3 & \hfil - \hfil & 17.8 & \hfil - \hfil \\
     \hline
     \multicolumn{4}{l}{(supervised keyphrase extraction)} \\
    Transformer & 110M & 28.8$_{5}$ & 42.7$_{5}$ & 25.1$_{9}$ & 48.7$_{11}$ \\
    Transformer+CRF & 110M & 28.2$_{6}$ & 43.2$_{3}$ & 26.0$_{8}$ & 52.0$_{11}$ \\
    BERT-base & 110M &  34.0$_{4}$ & 49.3$_{3}$ & 29.0$_{8}$ & 56.8$_{9}$ \\
    BERT-base+CRF & 110M &  33.9$_{7}$ & 49.9$_{6}$ & 28.5$_{5}$ & 56.3$_{5}$ \\
    SciBERT & 110M  & 31.8$_{3}$ & 47.7$_{2}$ & 29.2$_{5}$ & 57.5$_{5}$ \\
    SciBERT+CRF & 110M &  31.8$_{6}$ & 48.1$_{5}$ & 28.6$_{3}$ & 57.1$_{3}$ \\
    NewsBERT & 110M & 34.5$_{5}$ & 50.4$_{4}$ & 28.5$_{3}$ & 56.2$_{2}$ \\
    NewsBERT+CRF & 110M & 34.9$_{4}$ & 50.8$_{5}$ & 28.5$_{2}$ & 56.2$_{2}$ \\
    RoBERTa-base & 125M & 33.2$_{2}$ & 48.9$_{2}$ & 28.7$_{2}$ & 56.2$_{5}$ \\
    RoBERTa-base+CRF & 125M & 32.4$_{6}$ & 48.4$_{3}$ & 27.6$_{10}$ & 55.4$_{10}$ \\
    \hline
    \multicolumn{4}{l}{(supervised keyphrase generation)} \\
    CatSeq & 21M & 29.5 & 45.3 & \hfil - \hfil & \hfil - \hfil \\
    ExHiRD-h & 22M & 32.1$_{16}$ & 45.2$_{7}$ & 28.8$_{2}$ & 54.8$_{2}$ \\
    Transformer & 98M & 30.2$_{5}$ & 45.3$_{6}$ & 30.8$_{5}$ & 55.4$_{2}$ \\
    SetTrans & 98M & 35.6$_{5}$ & 46.3$_{4}$ & 35.8$_{3}$ & 56.7$_{5}$ \\
    \hdashline
    BERT-G & 110M & 32.3$_{5}$ & 47.4$_{3}$ & 28.5$_{6}$ & 54.9$_{6}$ \\
    SciBERT-G & 110M & 33.0$_{1}$ & 48.4$_{1}$ & 29.5$_{3}$ & 56.5$_{1}$ \\
    NewsBERT-G & 110M & 33.0$_{2}$ & 48.0$_{1}$ & 28.8$_{1}$ & 55.3$_{0}$ \\
    RoBERTa-G & 125M & 33.0$_{2}$ & 48.2$_{5}$ & \hfil - \hfil & \hfil - \hfil \\
    UniLM & 110M & 33.2$_{3}$ & 48.0$_{2}$ & \hfil - \hfil & \hfil - \hfil \\
    B2B-2+10 & 158M & 31.6$_{5}$ & 46.5$_{6}$ & 28.8$_{}$ & 55.4$_{}$ \\
    B2B-4+8 & 153M & 32.9$_{2}$ & 47.6$_{1}$ & 29.4$_{}$ & 55.7$_{}$ \\
    B2B-6+6 & 148M & 33.8$_{2}$ & 48.4$_{2}$ & 29.0$_{}$ & 55.6$_{}$ \\
    B2B-8+4 & 143M & 33.8$_{4}$ & 48.6$_{2}$ & 28.6$_{}$ & 55.6$_{}$ \\
    B2B-10+2 & 139M & 33.5$_{4}$ & 48.4$_{4}$ & 29.1$_{}$ & 55.9$_{}$ \\
    \hdashline
    BART-base & 140M & 35.9$_{1}$ & 49.9$_{2}$ & 30.4$_{1}$ & 57.1$_{1}$ \\
    BART-large & 406M & 37.3$_{16}$ & 51.0$_{15}$ & 31.2$_{2}$ & 57.8$_{8}$ \\
    T5-base & 223M & 34.6$_{2}$ & 49.2$_{2}$ & 28.7$_{1}$ & 56.1$_{1}$ \\
    T5-large & 770M & 36.6$_{0}$ & 50.8$_{1}$ & 30.5$_{2}$ & 58.0$_{3}$ \\
    KeyBART & 406M & 37.8$_{6}$ & 51.3$_{1}$ & 31.9$_{5}$ & 58.9$_{2}$ \\
    SciBART-base & 124M & 34.8$_{4}$ & 48.8$_{1}$ & 30.4$_{6}$ & 57.6$_{4}$ \\
    SciBART-large & 386M & 35.3$_{4}$ & 49.7$_{2}$ & 30.9$_{3}$ & 57.8$_{2}$ \\
    SciBART-large+OAGKX & 386M & 35.6$_{8}$ & 50.0$_{3}$ & 31.8$_{0}$ & 58.5$_{3}$ \\
    NewsBART-base & 140M & 35.4$_{2}$ & 49.8$_{1}$ & 30.7$_{3}$ & 57.5$_{0}$ \\
    \hline
    \end{tabular}
    \caption{Present keyphrase evaluation results of all the methods on KPTimes and StackEx. The reported results are averaged across three runs with different random seeds. The standard deviation of each entry is presented in the subscript. For example, 23.5$_6$ means an average of 23.5 with a standard deviation of 0.6. We omit the subscript for deterministic methods or methods with a single run.}
    \label{tab:other-all-results-pkp}
\end{table*}

\setlength{\tabcolsep}{4pt}
\begin{table*}[]
    \centering
    \begin{tabular}{l | c | l l | l l }
    \hline
    \multirow{2}{*}{Method} &
    \multirow{2}{*}{|M|} & \multicolumn{2}{c|}{\textbf{KPTimes}} & \multicolumn{2}{c}{\textbf{StackEx}} \\
     &  & F1@5 & F1@M & F1@5 & F1@M \\
    \hline
    CatSeq & 21M & 15.7 & 22.7 & \hfil - \hfil & \hfil - \hfil \\
    ExHiRD-h & 22M & 13.4$_{2}$ & 16.5$_{1}$ & 10.1$_{1}$ & 15.5$_{1}$ \\
    Transformer & 98M & 17.1$_{1}$ & 23.1$_{1}$ & 10.4$_{2}$ & 18.7$_{2}$ \\
    \hdashline
    SetTrans & 98M & 19.8$_{3}$ & 21.9$_{2}$ & 13.9$_{1}$ & 20.7$_{0}$ \\
    BERT-G & 110M & 16.5$_{7}$ & 24.6$_{5}$ & 10.7$_{8}$ & 22.8$_{10}$ \\
    SciBERT-G & 110M & 15.7$_{2}$ & 24.7$_{0}$ & 11.8$_{5}$ & 24.8$_{7}$ \\
    NewsBERT-G & 110M & 17.0$_{5}$ & 25.6$_{1}$ & 11.7$_{1}$ & 24.5$_{1}$ \\
    RoBERTa-G & 125M & 17.1$_{2}$ & 25.5$_{3}$ & \hfil - \hfil & \hfil - \hfil \\
    UniLM & 110M & 15.2$_{10}$ & 24.1$_{11}$ & \hfil - \hfil & \hfil - \hfil \\
    B2B-2+10 & 158M & 16.2$_{4}$ & 23.2$_{0}$ & 10.7$_{}$ & 22.3$_{}$ \\
    B2B-4+8 & 153M & 15.9$_{1}$ & 23.6$_{3}$ & 11.2$_{}$ & 22.8$_{}$ \\
    B2B-6+6 & 148M & 16.4$_{3}$ & 24.1$_{1}$ & 11.1$_{}$ & 22.9$_{}$ \\
    B2B-8+4 & 143M & 16.8$_{2}$ & 24.5$_{1}$ & 10.5$_{}$ & 22.2$_{}$ \\
    B2B-10+2 & 139M & 16.8$_{1}$ & 24.5$_{2}$ & 11.1$_{}$ & 23.4$_{}$ \\
    \hdashline
    BART-base & 140M & 17.1$_{2}$ & 24.9$_{1}$ & 11.7$_{0}$ & 24.9$_{2}$ \\
    BART-large & 406M & 17.6$_{10}$ & 24.4$_{19}$ & 12.4$_{1}$ & 26.1$_{3}$ \\
    T5-base & 223M & 15.3$_{1}$ & 24.2$_{1}$ & 9.4$_{0}$ & 21.6$_{1}$ \\
    T5-large & 770M & 15.7$_{1}$ & 24.1$_{1}$ & 10.6$_{1}$ & 23.9$_{2}$ \\
    KeyBART & 406M & 18.0$_{7}$ & 25.5$_{2}$ & 13.0$_{5}$ & 27.1$_{5}$ \\
    SciBART-base & 124M & 17.2$_{3}$ & 24.6$_{2}$ & 11.1$_{6}$ & 24.2$_{8}$ \\
    SciBART-large & 386M & 17.2$_{3}$ & 25.7$_{2}$ & 12.6$_{1}$ & 26.7$_{1}$ \\
    SciBART-large+OAGKX & 386M & 17.4$_{4}$ & 25.8$_{4}$ & 13.4$_{3}$ & 27.9$_{3}$ \\
    NewsBART-base & 140M & 17.6$_{3}$ & 26.1$_{1}$ & 12.1$_{3}$ & 25.7$_{4}$ \\
    \hline
    \end{tabular}
    \caption{Absent keyphrase evaluation results of all the methods on KPTimes and StackEx. The standard deviation of each entry is presented in the subscript. For example, 23.5$_6$ means an average of 23.5 with a standard deviation of 0.6. We omit the subscript for deterministic methods or methods with a single run.}
    \label{tab:other-all-results-akp}
\end{table*}

\section{Qualitative Results}
In Figure \ref{example-outputs-kp20k}, we present a few qualitative examples of KP20k from BART, T5, SciBERT, SciBART, and KeyBART. 

\begin{figure*}[h!]
\small
\centering
\begin{tabular}{p{0.98\linewidth}}
    \hline  
    \textbf{Title:} a review of \textcolor{blue}{design pattern} mining techniques . \\
    \textbf{Abstract:} the quality of a software system highly depends on its architectural design . high quality software systems typically apply expert design experience which has been captured as \textcolor{blue}{design patterns} . as demonstrated solutions to recurring problems , \textcolor{blue}{design patterns} help to reuse expert experience in software system design . they have been extensively applied in the industry . mining the instances of \textcolor{blue}{design patterns} from the source code of software systems can assist in the understanding of the systems and the process of re engineering them . more importantly , it also helps to trace back to the original design decisions , which are typically missing in legacy systems . this paper presents a review on current techniques and tools for mining \textcolor{blue}{design patterns} from source code or design of software systems . we classify different approaches and analyze their results in a comparative study . we also examine the disparity of the \textcolor{blue}{discovery} results of different approaches and analyze possible reasons with some insight . \\
    \hdashline
    \textbf{Ground Truth:} \textcolor{blue}{design pattern}, \textcolor{blue}{discovery}, \textcolor{blue}{reverse engineering}
\\
    \hdashline
    \textbf{BART:} unrelated scheduling, mechanism design, approximation algorithms  \\
    \textbf{T5:} \textcolor{blue}{design patterns}, software architecture, software design, software reuse  \\
    \textbf{SciBERT:} \textcolor{blue}{design patterns}, software systems, software engineering, software mining \\
    \textbf{SciBART:} design pattern mining, software system design, \textcolor{blue}{reverse engineering}, software reuse \\
    \textbf{KeyBART:} \textcolor{blue}{design patterns}, software architecture, software reuse, software quality, software maintenance \\
    \hline
    \hline  
    \textbf{Title:} stabilization of \textcolor{blue}{second order nonholonomic systems} in canonical chained form .  \\
    \textbf{Abstract:} stabilization of a class of \textcolor{blue}{second order nonholonomic systems} in canonical chained form is investigated in this paper . first , the models of two typical \textcolor{blue}{second order nonholonomic systems} , namely , a three link planar manipulator with the third joint unactuated , and a kinematic redundant manipulator with all joints free and driven by forces torques imposing on the end effector , are presented and converted to second order chained form by transformations of coordinate and input . a discontinuous control law is then proposed to stabilize all states of the system to the desired equilibrium point exponentially . computer simulation is given to show the effectiveness of the proposed controller . \\
    \hdashline
    \textbf{Ground Truth:} \textcolor{blue}{second order nonholonomic systems}, \textcolor{blue}{canonical second order chained form}, \textcolor{blue}{underactuated manipulator}, \textcolor{blue}{discontinuous coordinate transformation}, \textcolor{blue}{discontinuous stabilization}
\\
    \hdashline
    \textbf{BART:} stabilization, \textcolor{blue}{second order nonholonomic systems}, canonical chained form, discontinuous control law   \\
    \textbf{T5:} stabilization, \textcolor{blue}{second order nonholonomic system}, canonical chained form, discontinuous control \\
    \textbf{SciBERT:} stabilization, nonholonomic system, canonical chained form, redundant manipulator \\
    \textbf{SciBART:} stabilization, \textcolor{blue}{second order nonholonomic systems}, canonical chained form, discontinuous control law \\
    \textbf{KeyBART:} stabilization, \textcolor{blue}{second order nonholonomic systems}, discontinuous control law \\
    \hline
    \hline
    \textbf{Title:} characterizing output processes of e m e k [digit] queues . \\
    \textbf{Abstract:} our goal is to study which conditions of the output process of a queue preserve the increasing failure rate ( \textcolor{blue}{ifr} ) property in the interdeparture time . we found that the interdeparture time does not always preserve the \textcolor{blue}{ifr} property , even if the interarrival time and service time are both \textcolor{blue}{erlang distributions} with \textcolor{blue}{ifr} . we give a theoretical analysis and present numerical results of e m e k [digit] queues . we show , by numerical examples , that the interdeparture time of e m e k [digit] retains the \textcolor{blue}{ifr} property if m > k. ( c ) [digit] elsevier ltd. all rights reserved .\\
    \hdashline
    \textbf{Ground Truth:} \textcolor{blue}{ifr}, \textcolor{blue}{erlang distribution}, \textcolor{blue}{departure process}, \textcolor{blue}{ph g [digit]}, \textcolor{blue}{queueing theory} \\
    \hdashline
    \textbf{BART:} output process, increasing failure rate, interdeparture time, \textcolor{blue}{erlang distribution} \\
    \textbf{T5:} output process, increasing failure rate, \textcolor{blue}{erlang distribution} \\
    \textbf{SciBERT:} increasing failure rate, interdeparture time, \textcolor{blue}{erlang distribution}, output process of a queue \\
    \textbf{SciBART:} output process, increasing failure rate, interdeparture time, \textcolor{blue}{erlang distribution}, \textcolor{blue}{queueing theory} \\
    \textbf{KeyBART:} output process, increasing failure rate, interdeparture time, \textcolor{blue}{erlang distribution}, \textcolor{blue}{queueing theory} \\
    \hline
    \hline  
    \textbf{Title:} optimal tool selection for \textcolor{blue}{2.5 d milling} , part [digit] a \textcolor{blue}{solid modeling approach} for construction of the voronoi mountain \\
    \textbf{Abstract:} \textcolor{blue}{cutter selection} is a critical subtask of machining process planning . in this two part series , we develop a robust approach for the selection of an optimal set of milling cutters for a 2.5 d generalized pocket . in the first article ( part [digit] ) , we present a \textcolor{blue}{solid modeling} approach for the construction of the \textcolor{blue}{voronoi mountain} for the pocket geometry , which is a 3d extension of the voronoi diagram . the major contributions of this work include ( [digit] ) the development of a robust and systematic procedure for construction of the \textcolor{blue}{voronoi mountain} for a multiply connected curvilinear polygon and ( b ) an extension of the \textcolor{blue}{voronoi mountain} concept to handle \textcolor{blue}{open edges} . \\
    \hdashline
    \textbf{Ground Truth:} \textcolor{blue}{2.5 d milling}, \textcolor{blue}{solid modelling}, \textcolor{blue}{voronoi mountain}, \textcolor{blue}{cutter selection}, \textcolor{blue}{open edges} \\
    \hdashline
    \textbf{BART:} \textcolor{blue}{solid modeling}, \textcolor{blue}{voronoi mountain}, \textcolor{blue}{cutter selection}, 2.5 d generalized pocket, curve generation  \\
    \textbf{T5:} tool selection, \textcolor{blue}{2.5 d milling}, voronoi diagram, machining geometry\\
    \textbf{SciBERT:} tool selection, milling, \textcolor{blue}{voronoi mountain}, pocket geometry, cutter path planning \\
    \textbf{SciBART:} tool selection, \textcolor{blue}{2.5 d milling}, \textcolor{blue}{voronoi mountain}, \textcolor{blue}{cutter selection}, voronoi diagram, pocket milling \\
    \textbf{KeyBART:} tool selection, \textcolor{blue}{2.5 d milling}, \textcolor{blue}{voronoi mountain}, process planning, generalized pocket, vlsi cad cam \\
    \hline
\end{tabular}
\caption{Example outputs from various PLMs on the SciKP benchmarks. Correct keyphrases are colored in \textcolor{blue}{blue}.}
\label{example-outputs-kp20k}
\end{figure*}

\end{document}